\documentclass[10pt,twocolumn,letterpaper]{article}

\usepackage{iccv}
\usepackage{times}
\usepackage{epsfig}
\usepackage{graphicx}
\usepackage{amsmath}
\usepackage{amssymb}

\usepackage{amsmath,amsfonts,bm}

\def\eqref#1{Eq.~\ref{#1}}

\def\1{\bm{1}}

\def\ve{{\bm{e}}}

\def\vh{{\bm{h}}}

\def\vy{{\bm{y}}}

\DeclareMathAlphabet{\mathsfit}{\encodingdefault}{\sfdefault}{m}{sl}
\SetMathAlphabet{\mathsfit}{bold}{\encodingdefault}{\sfdefault}{bx}{n}

\usepackage{url}
\usepackage{amsthm}
\usepackage[utf8]{inputenc} 
\usepackage[T1]{fontenc}    
\usepackage{booktabs}       
\usepackage{etoolbox}
\usepackage{siunitx}        
\robustify\bfseries
\usepackage{amsfonts}       
\usepackage{nicefrac}       
\usepackage{microtype}      
\usepackage{bm}
\usepackage{color}
\usepackage{extarrows}
\usepackage{subfigure}
\usepackage{multirow}
\usepackage{mathtools}
\usepackage{bbm}

\newcommand{\tdash}{\multicolumn{1}{c}{-}}

\newcommand{\vzeta}{\bm{\zeta}}

\newcommand{\fullname}{Audio Visual Scene Graph Segmenter\xspace}
\newcommand{\name}{AVSGS\xspace}
\newcommand{\ACdataset}{ASIW\xspace}
\newcommand{\ACfulldatasetname}{Audio Separation in the Wild\xspace}

\newcommand{\vid}{V}
\newcommand{\src}{\mathbf{s}}
\newcommand{\srcspec}{\mathbf{S}}
\newcommand{\nodes}{\mathcal{V}}
\newcommand{\node}{v}
\newcommand{\edge}{e}
\newcommand{\edges}{\mathcal{E}}
\newcommand{\graph}{\mathcal{G}}
\newcommand{\sgraph}{g}
\newcommand{\pobjclasses}{\mathcal{P}}
\newcommand{\pobjs}{P}
\newcommand{\pobj}{p}
\newcommand{\objclasses}{\mathcal{C}}

\newcommand{\loss}{\mathcal{L}}

\newcommand{\feat}{F}
\newcommand{\embed}{\vy}
\newcommand{\embedset}{Y}
\newcommand{\mspec}{\bf{X}}
\newcommand{\mask}{\bf{M}}

\newcommand{\reals}[1]{\mathbb{R}^{#1}}

\newcommand{\set}[1]{\left\{#1\right\}}

\DeclareMathOperator{\frcnn}{FRCNN}
\DeclareMathOperator{\gru}{GRU}

\usepackage[pagebackref=true,breaklinks=true,letterpaper=true,colorlinks,bookmarks=false]{hyperref}

\iccvfinalcopy 

\def\iccvPaperID{10834} 

\ificcvfinal\fi

\begin{document}

\title{Visual Scene Graphs for Audio Source Separation}

\author{Moitreya Chatterjee$^{1}$ \qquad Jonathan Le Roux$^{2}$ \qquad Narendra Ahuja$^{1}$ \qquad Anoop Cherian$^{2}\thanks{Corresponding author.}$\\
$^1$University of Illinois at Urbana-Champaign,
Champaign, IL 61820, USA\\
$^2$Mitsubishi Electric Research Laboratories,
Cambridge, MA 02139, USA\\
\small{\texttt{metro.smiles@gmail.com\quad leroux@merl.com\quad n-ahuja@illinois.edu\quad cherian@merl.com}}
}

\maketitle
\ificcvfinal\thispagestyle{empty}\fi

\begin{abstract}
State-of-the-art approaches for visually-guided audio source separation typically assume sources that have characteristic sounds, such as musical instruments. These approaches often ignore the visual context of these sound sources or avoid modeling object interactions that may be useful to better characterize the sources, especially when the same object class may produce varied sounds from distinct interactions. To address this challenging problem, we propose \emph{\fullname}~(\name), a novel deep learning model that embeds the visual structure of the scene as a graph and segments this graph into subgraphs, each subgraph being associated with a unique sound obtained by co-segmenting the audio spectrogram. At its core, \name\ uses a recursive neural network that emits mutually-orthogonal sub-graph embeddings of the visual graph using multi-head attention. These embeddings are used for conditioning an audio encoder-decoder towards source separation. Our pipeline is trained end-to-end via a self-supervised task consisting of separating audio sources using the visual graph from artificially mixed sounds. 

In this paper, we also introduce an ``in the wild'' video dataset for sound source separation that contains multiple non-musical sources, which we call~\emph{\ACfulldatasetname}~(\ACdataset). This dataset is adapted from the AudioCaps dataset, and provides a challenging, natural, and daily-life setting for source separation. Thorough experiments on the proposed \ACdataset\ and the standard MUSIC datasets demonstrate state-of-the-art sound separation performance of our method against recent prior approaches. 
\end{abstract}

\section{Introduction}
 Real-world events often encompass spatio-temporal interactions of objects, the signatures of which leave imprints both in the visual and auditory domains when captured as videos. Knowledge of these objects and the sounds that they produce in their natural contexts are essential when designing artificial intelligence systems to produce meaningful deductions. For example, the sound of a cell phone \emph{ringing} is drastically different from that of one \emph{dropping on the floor}; such distinct sounds of objects and their contextual interactions may be essential for an automated agent to assess the scene. The importance of having algorithms with such audio-visual capabilities is far reaching, with applications such as 
 audio denoising, musical instrument equalization, audio-guided visual surveillance, or even in navigation planning for autonomous cars, for example by  \emph{visually localizing the sound of an ambulance}. 
 
 \begin{figure}[t]
    \centering
    \includegraphics[width=8cm,trim={2cm 8cm 2cm 2cm},clip]{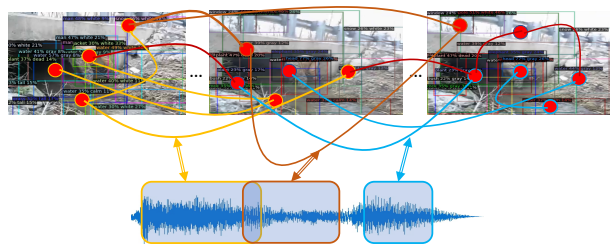}
   \caption{A schematic illustration of our \fullname (\name) framework on frames from our Audio Separation in the Wild (ASIW) dataset. Given an input video and the associated audio, our method builds a spatio-temporal (fully-connected) visual scene graph spanning across the video frames, and learns alignments between sub-graphs of this scene graph and the respective audio regions. The use of the scene graph allows rich characterization of the objects and their interactions, allowing effective identification of the sound sources for better separation. }
    \label{fig:setup_fig}
    \vspace{-0.2cm}
\end{figure}

 Recent years have seen a surge in algorithms at the intersection of visual and auditory domains, among which visually-guided source separation -- the problem of separating sounds from a mixture using visual cues -- has made significant strides~\cite{gan2020music,gao2019co,zhao2018sound,zhao2019sound}. State-of-the-art algorithms for this task~\cite{xu2019recursive,zhao2018sound,zhao2019sound} typically restrict the model design to only objects with unique sounds (such as musical instruments~\cite{gan2020music, zhao2019sound}) or consider settings where there is only a single sound source, and the models typically lack the richness to capture spatio-temporal audio-visual context.  For example, for a video with \emph{``a guitar being played by a person''} and one in which \emph{``a guitar is kept against a wall''}, the context may help a sound separation algorithm to decide whether to look for the sound of the guitar in the audio spectrogram; however several of the prior works only consider visual patches of an instrument as the context to guide the separation algorithm~\cite{gao2019co}, which is sub-optimal. 
 
 From a learning perspective, the problem of audio-visual sound source separation brings in several interesting challenges: (i) The association of a visual embedding of a sound source to its corresponding audio can be a one-to-many mapping and therefore ill-posed. For example, \emph{a dog barking while splashing water in a puddle}. Thus, methods such as~\cite{gan2020music,gao2019co} that assume a single visual source may be misled. (ii) It is desirable that algorithms for source separation are scalable to new sounds and their visual associations; i.e., the algorithm should be able to master the sounds of varied objects (unlike supervised approaches~\cite{wang2018supervised,xia2017using}). (iii) Naturally occurring sounds can emanate out of a multitude of interactions -- therefore, using \emph{a priori} defined sources, as in~\cite{gan2020music,gao2019co}, can be limiting.

In this work, we rise up to the above challenges using our \fullname\ (\name) framework for the concrete task of sound source separation. Figure~\ref{fig:setup_fig} presents the input-output setting for our task. Our setup represents the visual scene using spatio-temporal scene graphs~\cite{johnson2015image} capturing visual associations between objects occurring in the video, towards the goal of training \name~ to infer which of these visual associations lead to auditory grounding. To this end, we design a recursive source separation algorithm (implemented using a GRU) that, at each recurrence, produces an embedding of a sub-graph of the visual scene graph using graph multi-head attention. These embeddings are then used as conditioning information to an audio separation network, which adopts a U-Net style encoder-decoder architecture~\cite{ronneberger2015u}. As these embeddings are expected to \emph{uniquely} identify a \emph{sounding interaction}, we enforce that they be mutually orthogonal. We train this system using a self-supervised approach similar to Gao \etal \cite{gao2019co}, 
wherein the model is encouraged to disentangle the audio corresponding to the conditioned visual embedding from a mixture of two or more different video sounds. Importantly, our model is trained to ensure consistency of each of the separated sounds by their type, across videos. Thus, two guitar sounds from two disparate videos should sound more similar than a guitar and a piano. Post separation, the separated audio may be associated with the visual sub-graph that induced its creation, making the sub-graph an \textit{Audio-Visual Scene Graph} (AVSG), usable for other downstream tasks.

We empirically validate the efficacy of our method on the popular Multimodal Sources of Instrument Combinations (MUSIC) dataset~\cite{zhao2018sound} and a newly adapted version of the AudioCaps dataset~\cite{kim2019audiocaps}, which we call \ACfulldatasetname (\ACdataset). The former contains videos of performers playing musical instruments, while the latter features videos of naturally occurring sounds arising out of complex interactions \textit{in the wild}, collected from YouTube. Our experiments demonstrate the importance of visual context in sound separation, and \name outperforms prior state-of-the-art methods on both of these benchmarks.

We now summarize the key contributions of the paper:
\begin{itemize}\setlength\itemsep{-0.2em}
    \item To the best of our knowledge, ours is the first work to employ the powerful scene graph representation~\cite{johnson2015image} for the task of visually-guided audio source separation.
	\item We present \name~for this task, that is trained to produce mutually-orthogonal embeddings of the visual sub-graphs, allowing our model to infer representations of \textit{sounding interactions} in a self-supervised way.
	\item We present \ACdataset, a large scale \textit{in the wild} dataset adapted from \textit{AudioCaps} for the source separation task. This dataset features sounds arising out of natural and complex interactions.
	\item Our \name ~framework demonstrates state-of-the-art performance on both the datasets for our task.
\end{itemize}

\section{Related Works}
In this section, we review relevant prior works which we group into several categories for ease of readability.

\noindent \textbf{Audio Source Separation} has a very long history in the fields of signal processing and more recently machine learning \cite{comon2010handbook,gannot2017consolidated,loizou2013speech,vincent2018audio,wang2018supervised,wang2006computational}. \textit{Audio-only} methods have typically either relied on \textit{a priori} assumptions on the statistics of the target sounds (such as independence, sparsity, etc.), or resorted to supervised training to learn these statistics~\cite{smaragdis2014static} (and/or optimize the separation process from data~\cite{weninger2014nmf}) via deep learning~\cite{xu2013experimental,huang2014deep,Weninger2014GlobalSIP12,Hershey2016ICASSP03,yu2017permutation,wang2018supervised}. Such supervised learning often involves the creation of synthetic training data by mixing known sounds together and training the model to recover target sounds from the mixture. Settings where isolated target sources are unavailable have recently been considered, either by relying on weak sound-event activity labels \cite{pishdadian2020finding}, or using unsupervised methods that learn to separate mixtures of mixtures \cite{wisdom2020unsupervised}. 

\noindent \textbf{Audio-Visual Source Separation} considers the task of discovering the association between the acoustic signal and its corresponding signature in the visual domain. 
Such methods have been employed for tasks like speech separation \cite{Afouras20b,ephrat2018looking,michelsanti2020overview}, musical instrument sound separation \cite{gao2019co,gan2020music,zhao2019sound,zhao2018sound}, and separation of on-screen sounds of generic objects \cite{owens2018audio,tzinis2020into}. More recently, researchers have sought to integrate motion information into the visual representation of these methods, either in the form of pixel trajectories \cite{zhao2019sound}, or human pose \cite{gan2020music}. However,
these approaches adopt a video-level ``mix-and-separate'' training strategy which works best with clean, single-source videos. Differently, our approach is trained to disentangle sound sources within a video. Gao \etal ~\cite{gao2019co} proposed an approach in a similar regime, however they do not capture the visual context, which may be essential to separate sound that emanates as a result of potentially complex interactions between objects in the scene. Further, our proposed framework allows characterizing generic sounds that can arise from fairly unconstrained settings, unlike approaches that are tailored to tasks such as musical instrument sound separation.

\noindent \textbf{Localizing Sound in Video Frames} seeks to identify the pixels in a video frame that visually represent the sound source. Several approaches have been proposed for this task~\cite{arandjelovic2018objects,kidron2005pixels,senocak2018learning,kidron2005pixels,hershey1999audio}. While such methods do visually ground the audio sources, they do not separate the audio, which is the task we consider. 

\noindent \textbf{Synthesizing Sound from Videos} constitutes another class of techniques in the audio-visual paradigm~\cite{owens2016visually,zhou2018visual} that has become popular in recent years. For example,~\cite{gao20192,morgado2018self} propose frameworks capable of generating both monaural and binaural audio starting from videos. However, we are interested in separating the audio from different sound sources, starting with a mixed audio.

\noindent \textbf{Scene Graphs in Videos} have proven to be an effective toolkit in representing the content of static images~\cite{johnson2015image,li2008modeling} capable of capturing the relationship between different objects in the scene. These representations have only recently been deployed to videos for tasks such as action recognition~\cite{ji2020action} and  visual dialog~\cite{geng2020spatio}. We employ these powerful representations to separate a mixed audio into its constituent sources, which can then be associated with their corresponding sub-graphs for other downstream tasks.

\section{Proposed Method}
We begin this section by first presenting a description of the problem setup along with an overview of our model. We then delve deeper into the details of the model and finish the section by providing the details of our training setup.

\begin{figure*}[t]
    \centering
    \includegraphics[width=12cm,trim={0cm 4cm 1.5cm 3cm},clip]{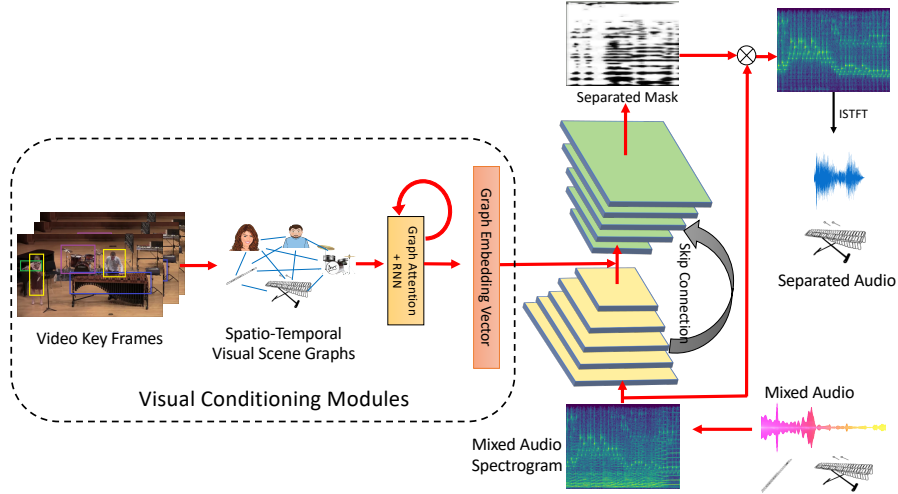}
   \caption{Detailed illustration of our proposed \name model. }
    \label{fig:arch}
    \vspace{-0.3cm}
\end{figure*}

\subsection{Problem Setup and Overview}
Given an unlabeled video $\vid$ and its associated discrete-time audio $x(t) = \sum_{i=1}^N\src_{i}(t)$ consisting of a linear mixture of $N$ audio sources $\src_{i}(t)$, the objective in \textit{visually-guided} source separation is to use $\vid$ to disentangle $x(t)$ into its constituent sound sources $\src_{i}(t)$, for $i \in \{1,2, \dots, N\}$. In this work, we represent the video as a spatio-temporal visual scene graph $\graph = (\nodes, \edges)$, with nodes $\nodes = \set{\node_1, \node_2, \dots, \node_K}$ representing objects (including people) in the video $\vid$, and $\edges$ denoting the set of edges $\edge_{jk}$ capturing the pairwise interaction or spatial context between nodes $\node_j$ and $\node_k$. Our main idea in \name is to learn to associate each audio source $\src_i(t)$ with a visual sub-graph $\sgraph_i$ of $\graph$. We approach this problem from the perspective of graph attention pooling to produce mutually-orthogonal sub-graph embeddings auto-regressively; these embeddings are made to be aligned with the respective audio sources using an \emph{Audio Separator} sub-network that is trained against a self-supervised \emph{unmixing} task~\cite{gao2019co,zhao2018sound,zhao2019sound}. Figure~\ref{fig:arch} presents an overview of the algorithmic pipeline of our model.

\subsection{\fullname Model}
Figure~\ref{fig:arch} presents an illustration of the algorithmic pipeline that we follow in order to obtain the separated sounds $\src_{i}(t)$ from their mixture $x(t)$. Below, we present the details of each step of this pipeline.

\noindent\textbf{Object Detector:}
The process of representing a video $\vid$ as a spatio-temporal scene graph starts with detecting a set of $M$ objects and their spatial bounding boxes in each frame of the video. As is common practice, we use a Faster-RCNN ($\frcnn$)~\cite{ren2016faster} model for this task, trained on the Visual Genome dataset~\cite{krishna2017visual}. As this dataset provides around 1600 object classes, denoted $\objclasses$, it can detect a significant set of common place objects. Further, for detecting objects that are not in the Visual Genome classes (for example, the musical instruments in the MUSIC dataset which we consider later), we trained a separate $\frcnn$ model with labeled images from the Open Images dataset~\cite{krasin2017openimages}, which contains those instrument annotations.    

Given a video frame $I$, the object detector $\frcnn$ produces a set of $M$ quadruples $\set{\left(C^{k}_{I}, B^{k}_{I}, \feat^{k}_{I}, S^{k}_{I}\right)}_{k=1}^M = \frcnn(I)$, one for each detected object, consisting of the label  $C\in\objclasses$ of the detected object, its bounding box $B$ in the frame, a feature vector $F$ identifying the object, and a detection confidence score $S$.

\noindent \textbf{Visual Scene Graph Construction}: 
Once we have the object detections and their meta-data, our next sub-task is to use this information to construct our visual scene graph. While standard scene graph approaches~\cite{ji2020action} often directly use the object detections to build the graph (sometimes combined with a visual relationship detector~\cite{geng2020spatio}), our task of sound separation demands that the graph be constructed in adherence to the audio, so that the audio-visual correlations can be effectively learned. To this end, for every sound of interest, we associate a \emph{principal object}, denoted $\pobj$, among the classes in $\objclasses$ (obtainable from the $\frcnn$) that could have produced the sound. For example, for the sound of a \emph{piano} in an orchestra, the principal object can be the piano, while for the sound of \emph{ringing}, the object could be a \emph{telephone}. Let us denote the set of such principal object classes as $\pobjclasses\subset\objclasses$.

To construct the visual scene graph for a given video $\vid$, we first identify the subset of principal objects $\pobjs= \{\pobj_{1}, \dots, \pobj_{{N}}\}\subset\pobjclasses$ that are associated with that video. This information is derived from the video metadata, such as for example the video captions or the class labels, if available. Next, we identify the video frames containing the most confident detections of each object $\pobj_{i}\in\pobjs$. We refer to such frames as the \textit{key frames} of the video -- our scene graph is constructed using these key frames. For every principal object $\pobj_{i}$, we then identify the subset of the $M$ object bounding boxes (produced by FRCNN for that key frame), which have an Intersection Over Union (IoU) with the bounding box for $\pobj_{i}$ greater than a pre-defined threshold $\gamma$. We refer to this overlapping set of nodes as the \textit{context nodes} of $\pobj_{i}$, denoted as $\nodes_{\pobj_{i}}$. The vertex set of the scene graph is then constructed as $\nodes = \bigcup_{i=1}^{N} (\pobj_{i} \bigcup  \nodes_{\pobj_{i}})$. Note that each graph node $\node$ is associated with a feature vector $\feat_\node$ produced by $\frcnn$ for the visual patch within the respective bounding box.
 
Our next sub-task for scene graph construction is to define the graph edges $\edges$. Due to the absence of any supervision to select the edges (and rather than resorting to heuristics), we assume useful edges will emerge automatically from the audio-visual co-segmentation task, and thus, we decided to use a fully-connected graph between all the nodes in $\nodes$; i.e., our edges are given by: $\edges=\set{\edge_{jk}}_{(j,k)\in\nodes\times\nodes}$. Since the scene graph is derived from multiple key frames in the video and its vertices span a multitude of objects in the key frames, our overall scene graph is thus spatio-temporal in nature.

\noindent \textbf{Visual Embeddings of Sounding Interactions:} The visual scene graph $\graph$ obtained in the previous step is a holistic representation of the video, and thus characterizes the visual counterpart of the mixed audio $x(t)$. To separate the audio sources from the mixture, \name must produce visual cues that can distinctly identify the sound sources. However, we neither know the sources nor do we know what part of the visual graph is producing the sounds. To resolve this dichotomy, we propose a joint learning framework in which the visual scene graph is segmented into sub-graphs, where each sub-graph is expected to be associated with a unique sound in the audio spectrogram, thus achieving source separation. To guide the model to learn to correctly achieve the audio-visual segmentation, we use a self-supervised task described in the next section. For now, let us focus on the modules needed to produce embeddings for the visual sub-graph.

For audio separation, there are two key aspects of the visual scene graph that we expect the ensuing embedding to encompass: (i) the nodes corresponding to sound sources and (ii) edges corresponding to sounding interactions. For the former, we use a multi-headed graph attention network~\cite{velivckovic2017graph}, taking as input the features $\feat_\node$ associated with the scene graph nodes $\node$ and implement multi-head graph message passing, thereby parting attention weights to nodes that the framework (eventually) learns to be important in characterizing the sound. For the latter, i.e., capturing the interactions, we design an edge convolution network~\cite{wang2019dynamic}. These networks are typically multi-layer perceptrons, $\vh_{\Lambda}(\cdot, \cdot)$, which take as input the concatenated features corresponding to a pair of nodes $\node_j$ and $\node_k$ which are connected by an edge and produces an output vector $\ve_{jk}$. $\Lambda$ encapsulates the learnable parameters of this layer. The updated features of a node $\node_k$ are then obtained by averaging all the edge convolution embeddings incident on $\node_k$. The two modules are implemented in a cascade with the node attention preceding the edge convolutions. Next, the attended scene graph is pooled using global max-pooling and global average pooling~\cite{lee2019self}; the pooled features from each operation are then concatenated, resulting in an embedding vector $\vzeta$ for the entire graph. As we need to produce $N$ embedding vectors from $\vzeta$, one for each source and another additional one for background, we need to keep track of the embeddings generated thus far. To this end, we propose to use a recurrent neural network, implemented using a GRU. In more detail, our final set of visual sub-graph embeddings $\embedset=\set{\vy_1, \vy_2,\dots, \vy_N, , \vy_{N+1}}$, where each $\embed_i\in\reals{d}$, is produced auto-regressively as:
\begin{equation}
    \embed_i = \gru(\vzeta; \Delta_{i-1}),\ i=1,2,\dots, N, N+1
\end{equation}
where $\Delta_{i-1}$ captures the bookkeeping that the GRU does to keep track of the embeddings generated thus far. 

\noindent\textbf{Mutual-Orthogonality of Visual Embeddings:} 
A subtle but important technicality that needs to be addressed for the above framework to succeed is in allowing the GRU to know whether it has generated embeddings for all the audio sources in the mixture. This poses the question \emph{how do we ensure the GRU does not repeat the embeddings?} Practically, we found that this is an important ingredient in our setup for audio source separation. To this end, we propose to enforce mutual orthogonality between the embeddings that the GRU produces. That is, for each recurrence of the GRU, it is expected to produce a unit-normalized embedding $\embed_i$ that is orthogonal to each of the embeddings generated prior to it, i.e., $\set{\embed_1, \embed_2,\dots, \embed_{i-1}}$. We include this constraint as a regularization in our training setup. Mathematically, we enforce a softer-version of this constraint given by:
\begin{align}
    \loss_{\mathrm{ortho}}(\embedset) = \sum_{i, j \in \{1,2, \dots, N\},  i \neq j} (\embed_i^\top {\embed}_j) ^{2}.
    \label{eq:dotpdt}
\end{align}
One key attribute of this mechanism for deriving the feature representations $\embed_{i}$ is that such embeddings could emerge from potentially complex interactions between the objects in the scene graph, unlike popular prior approaches, which resort to more simplistic visual embeddings, such as the whole frame~\cite{zhao2019sound} or a single object~\cite{gao2019co}.    

\noindent \textbf{Audio Separator Network:} The final component in our model is the \textit{Audio Separator Network} (ASN). Given the success of U-Net \cite{ronneberger2015u} style encoder-decoder networks for separating audio mixtures into their component sound sources \cite{jansson2017singing,liu2019divide}, particularly in conditioned settings \cite{gao2019co,meseguer2019conditioned,zhao2018sound,slizovskaia2019end}, we adopt this architecture for inducing the source separation. Since we are interested in \textit{visually guiding} the source separation, we condition the bottleneck layer of ASN with the sub-graph embeddings $\embed_{i}$ produced above. In detail, ASN takes as input the magnitude spectrogram ${\mspec}\in\reals{\Omega\times T}$ of a mixed audio $x(t)$, produced via the short-time Fourier transform (STFT), where $\Omega$ and $T$ denote the number of frequency bins and the number of video frames, respectively. The spectrogram is passed through a series of 2D-convolution layers, each coupled with Batch Normalization and Leaky ReLU, until we reach the bottleneck layer. At this layer, we replicate each graph embedding $\embed_{i}$ to match the spatial resolution of the U-Net bottleneck features, and concatenate along its channel dimension. This concatenated feature tensor is then fed to the U-Net decoder. The decoder consists of a series of up-convolution layers, followed by non-linear activations, each coupled with a skip connection from a corresponding layer in the U-Net encoder and matching in spatial resolution of its output. The final output of the U-Net decoder is a time-frequency mask, $\hat{\mask}_{i} \in [0, 1]^{\Omega \times T}$, which when multiplied with the magnitude spectrogram $\mspec$ of the mixture yields an estimate of the magnitude spectrogram of the separated source $\hat{\srcspec}_{i}= \hat{\mask}_{i} \odot {\mspec}$, where $\odot$ denotes element-wise product. An estimate $\hat{\src}_{i}(t)$ of the separated waveform signal for the $i$-th source can finally be obtained by applying an inverse short-time Fourier transform (iSTFT) to the complex spectrogram obtained by combining $\hat{\srcspec}_{i}$ with the mixture phase. 
For architectural details, please refer to the supplementary.

\subsection{Training Regime}
Audio source separation networks are typically trained in a supervised setting in which a synthetic mixture is created by mixing multiple sound sources including one or more \emph{known} target sounds, and training the network to estimate the target sounds when given the mixture as input \cite{huang2014deep,Hershey2016ICASSP03,wang2018supervised,Weninger2014GlobalSIP12,xu2013experimental,yu2017permutation}. In the visually-guided source separation paradigm, building such synthetic data by considering multiple videos and mixing their sounds is referred to as ``mix-and-separate''~\cite{gao2019co,gan2020music,zhao2019sound,zhao2018sound}.  We train our model in a similar fashion to Gao \etal \cite{gao2019co}, in which a co-separation loss is introduced to allow separation of multiple sources within a video without requiring ground-truth signals on the individual sources.

In this training regime, we feed the ASN with a spectrogram representation $\mspec_{\mathrm{m}}$ of the mixture $x_{\mathrm{m}}(t) = x_{1}(t) + x_{2}(t)$ of the audio tracks from two videos, and build representative scene graphs, $\graph_1$ and  $\graph_2$, for each of the two corresponding videos. We then extract unit-norm embeddings from each of these two scene graphs, $\embed^{1}_{i}, i \in \{1, 2, \dots, N_1\}$ and $\embed^{2}_{i}, i \in \{1, 2, \dots, N_2\}$. Next, each of these embeddings $\embed^{u}_{i}$ are independently pushed into the bottleneck layer of ASN that takes as input $\mspec_{\mathrm{m}}$. Once a separated spectrogram $\hat{\srcspec}^{u}_{i}$ is obtained as output for the input pair $(\embed^{u}_{i}, \mspec_{\mathrm{m}})$, we feed this $\hat{\srcspec}^{u}_{i}$ to a classifier which enforces the spectrogram signature to be classified as that belonging to one of the principal object classes in $\pobjs_u$. 
In contrast with \cite{gao2019co}, where there is a direct relationship between the conditioning by a visual object and the category of the sound to be separated, we here do not know in which order the GRU produced the conditioning embeddings, and thus which principal object class $l^u_c \in \pobjs_u$ should correspond to a given embedding $\embed^{u}_{i}$.

We therefore consider different permutations $\sigma^u$ of the ground-truth class labels of video $u$, matching the ground-truth label of the $c$-th object to the $\sigma^u(c)$-th embedding, and use the one which yields the minimum cross-entropy loss, similarly to the permutation free (or invariant) training employed in speech separation \cite{Hershey2016ICASSP03,Isik2016,yu2017permutation}. Our loss is then:
\begin{equation}
\mathcal{L}_{\mathrm{cons}} = - \sum_{u=1,2} \min_{\sigma^u\in \mathcal{S}_{N_u+1}} 
\sum_{c=1}^{N_u+1}   \log (p^u_{\sigma^u(c)}(l^u_c)),
\label{eq:consistency}
\end{equation}
where $\mathcal{S}_{N_u+1}$ indicates the set of all permutations on $\{1,\dots,N_u+1\}$, $p^u_i(l)$ denotes the predicted probability produced by the classifier for class $l$ given $\hat{\srcspec}^{u}_{i}$ as input, and $l^u_c$ is the ground-truth class of the $c$-th object in video $u$. 

Further, in order to restrict the space of plausible audio-visual alignments and to encourage the ASN to recover full sound signals from the mixture (in contrast to merely what is required to minimize the consistency loss~\cite{pishdadian2020finding}), we also ensure that the sum of the predicted masks for separating the sound sources produce an estimated mask that is close to the ground truth ideal binary mask \cite{li2009optimality}, using a co-separation loss similar to prior work \cite{gao2019co,pishdadian2020finding}:
\begin{equation}
\mathcal{L}_{\mathrm{co-sep}} = \sum_{u=1,2} \Big\|\sum_{i=1}^{N_u + 1} \hat{\mask}^{u}_i - \mask^u_{\mathrm{ibm}}\Big\|_1,
\label{eq:cyclic}
\end{equation}
where $\mask^u_{\mathrm{ibm}}=\mathbbm{1}_{X_{\mathrm{m}}^u>X_{\mathrm{m}}^{\neg u}}$ denotes the ideal binary mask for the audio of video $u$ within the mixture $X_{\mathrm{m}}$. 

Armed with the above three losses in~\eqref{eq:dotpdt},~\eqref{eq:consistency}, and~\eqref{eq:cyclic}, the final training loss for our model is obtained as follows, with weights $\lambda_1, \lambda_2, \lambda_3 \geq 0$:
\begin{equation}
\mathcal{L} = \lambda_1 \mathcal{L}_{\mathrm{cons}} + \lambda_2 \mathcal{L}_{\mathrm{co-sep}} + \lambda_3 \mathcal{L}_{\mathrm{ortho}}.
\end{equation}

\section{Experiments}
In order to validate the efficacy of our approach, we conduct experiments on two challenging datasets and compare its performance against competing and recent baselines.

\begin{table*}[t]
 \centering
  \sisetup{table-format=2.1,round-mode=places,round-precision=1,table-number-alignment = center,detect-weight=true,detect-inline-weight=math}
 \caption{SDR, SIR, and SAR [dB] results on the MUSIC and ASIW test sets. 
 [Key: Best results in \textbf{bold} and second-best in \textcolor{blue}{blue}.] }\label{tab:perf_asiw_music}
 \begin{tabular}{lSSSSSS}
\toprule
\multirow{2}{*}{} & \multicolumn{3}{c}{{MUSIC}} & \multicolumn{3}{c}{{ASIW}} \\ 
\cmidrule(l{0.25em}r{0.25em}){2-4}\cmidrule(l{0.25em}r{0.25em}){5-7}
& {SDR} $\uparrow$ & {SIR} $\uparrow$ &	{SAR} $\uparrow$ & {SDR} $\uparrow$ & {SIR} $\uparrow$ &	{SAR} $\uparrow$ \\ 
\midrule
Sound of Pixel (SofP)~\cite{zhao2018sound} & 6.05	& 10.9 &	10.63 &	6.19 & 8.12 &   10.62 \\  
Minus-Plus Net (MP Net) \cite{xu2019recursive} & 7.00	& 14.39 &	10.21 &	2.96 &	7.66 &	9.42 \\ 
Sound of Motion (SofM)~\cite{zhao2019sound} & 8.24	& 14.55 &	\color{blue}13.15 &	\color{blue}6.73 & 9.36 &	11.1 \\ 
Co-Separation~\cite{gao2019co} & 7.4	& 13.82 &	10.6 &	6.56 &	\color{blue}12.89 &	\color{blue}12.64 \\ 
Music Gesture (MG)~\cite{gan2020music} & \color{blue}10.09	& \color{blue}15.66 &	12.92 &	\tdash &	\tdash &	\tdash \\ \midrule
AVSGS (Ours) & \bfseries 11.43 &	\bfseries 17.25 &	\bfseries 13.52 &	\bfseries 8.75 &\bfseries 14.06 	& 	\bfseries 12.95 \\ 
\bottomrule
\end{tabular}
\vspace{-0.1cm}
\end{table*}

\begin{table*}[t]
 \centering
 \caption{Number of videos for each of the principal object categories of ASIW dataset. }\label{tab:asiw_stats}
 {\setlength{\tabcolsep}{4pt}
 \begin{tabular}{cccccccccccccc}
\toprule
  {Baby}	& {Bell} &	{Birds} &	{Camera} &	{Clock} &	{Dogs} &	{Toilet} &	{Horse} &	{Man} &	{Sheep} &	{Telephone} &	{Trains} &	{Vehicle} &	{Water} \\ 
  \midrule
    1616	& 151 &	2887 &	913 &	658 &	1407 &	838 &	385 &	6210 &	710 &	222 &	141 &	779 &	378 \\ 
    \bottomrule
\end{tabular}
}
\vspace{-0.3cm}
\end{table*}

\begin{table}[t]
 \centering
 \sisetup{table-format=2.1,round-mode=places,round-precision=1,table-number-alignment = center,detect-weight=true,detect-inline-weight=math}
 \caption{SDR, SIR, and SAR [dB] results on the ASIW test set. [Key: Best results in \textbf{bold}.] }\label{tab:asiw_ablation}
 \resizebox{\linewidth}{!}{
 \begin{tabular}{rlSSS}
\toprule
\multirow{2}{*}{} & \multicolumn{3}{c}{{ASIW}} \\ \cmidrule(l{0.25em}r{0.25em}){3-5}
{Row}&&  {SDR} $\uparrow$ & {SIR} $\uparrow$ &	{SAR} $\uparrow$  \\ 
\midrule
 1 & AVSGS (Full) & \bfseries 8.75 &\bfseries 14.06	& 	\bfseries 12.95  \\ \midrule
2 & AVSGS - No orthogonality ($\lambda_3 = 0$) &  7.37	&  13.28 &	11.56  \\ 
3 & AVSGS - No multi-lab. ($\lambda_1 = 0$) &  6.4 &	11.2 &	11.7  \\ 
4 & AVSGS - No co-sep ($\lambda_2 = 0$) &  1.1 &	1.3 &	\bfseries 13.8  \\ 
5 & AVSGS - N=3 &  8.4 &	13.5 &	12.2  \\ 
6 & AVSGS - No Skip Conn. &  2.8 &	4.6 &	11.3  \\ 
7 & AVSGS - No GATConv &  6.5 &	11.6 &	11.8  \\ 
8 & AVSGS - No EdgeConv &  6.2  &	10.1 &	13.2  \\ 
9 & AVSGS - No GRU &  6.49	&  12.33 &	10.64  \\ 
\bottomrule
\end{tabular}
}
\vspace*{-0.5cm}
\end{table}

\subsection{Datasets}
\noindent \textbf{Audio Separation in the Wild (ASIW):} Most prior approaches in visually-guided sound source separation report performances solely in the setting of separating the sounds of musical instruments ~\cite{gan2020music,zhao2018sound,zhao2019sound}. Given musical instruments often have very characteristic sounds and most of the videos used for evaluating such algorithms often contain professional footages, they may not capture the generalizability of those methods to daily-life settings. While there have been recent efforts towards looking at more natural sounds \cite{xu2019recursive}, the categories of audio they consider are limited ($\sim$10 classes). Moreover, most of the videos contain only a single sound source of interest, making the alignment straightforward.
There are a few datasets that could be categorized as considering ``in the wild'' source separation, such as \cite{gao2018learning,tzinis2020into}, but they either only consider separating between on-screen and off-screen sounds~\cite{tzinis2020into}, or provide only limited information about the nature of sounds featured~\cite{gao2018learning}, making the task of learning the audio-visual associations challenging.

To fill this gap in the evaluation benchmarks between ``in the wild'' settings and those with very limited annotations,  we introduce a new dataset, called \textit{Audio Separation in the Wild (ASIW)}. ASIW is adapted from the recently introduced large-scale AudioCaps dataset \cite{kim2019audiocaps}, which contains 49,838 training, 495 validation, and 975 test videos crawled from the AudioSet dataset \cite{gemmeke2017audio}, each of which is around 10 s long. In contrast to~\cite{gao2018learning}, these videos have been carefully annotated with human-written captions (English-speaking Amazon Mechanical Turkers -- AMTs), emphasizing the auditory events in the video. We manually construct a dictionary of 306 frequently occurring \textit{auditory words} from these captions. A few of our classes include: \textit{splashing}, \textit{flushing}, \textit{eruptions}, or \textit{giggling,} and these classes are almost always grounded to principal objects in the video generating the respective sound. The set of principal objects has 14 classes (baby, bell, birds, camera, clock, dogs, toilet, horse, man/woman, sheep/goat, telephone, trains, vehicle/car/truck, water) and an additional background class. The principal object list is drawn from the Visual Genome \cite{krishna2017visual} classes. We retain only those videos which contain at least one of these 306 auditory words. Table~\ref{tab:asiw_stats} gives a distribution of the number of videos corresponding to each of these principal object categories. The resulting dataset features audio both arising out of standalone objects, such as \emph{giggling of a baby}, as well as from inter-object interactions, such as \emph{flushing of a toilet by a human}. The supplementary material lists all the 306 auditory words and the principal object associated to each word. After pre-processing this list, we use 147 validation and 322 test videos in our evaluation, while 10,540 videos are used for training. 

\noindent \textbf{MUSIC Dataset:} Apart from our new ASIW dataset, we also report performance of our approach on the MUSIC dataset~\cite{zhao2018sound} which is often considered as the standard benchmark for visually-guided sound source seapartion. This dataset consists of 685 videos featuring humans performing musical solos and duets using 11 different instruments; 536 of these videos feature musical solos while the rest are duet videos. The instruments being played feature significant diversity in their type (for instance, guitar, erhu, violin are string instruments, flute, saxophone, trumpet are wind instruments, while xylophone is a percussion instrument). This makes the dataset a challenging one, despite its somewhat constrained nature. In order to conduct experiments, we split these videos into 10-second clips, following the standard protocol~\cite{gao2019co}. We ignore the first 10 seconds window of each of the untrimmed videos while constructing the dataset, since quite often the players do not really start playing their instruments right away. This results in 6,300/132/158 training, validation, and test videos respectively.

\subsection{Baselines} 
We compare \name against recently published approaches for visually-guided source separation, namely:\\
\noindent \textbf{Sound of Pixel (SofP)}~\cite{zhao2018sound}: one of the earliest deep learning based methods for this task.\\
\noindent \textbf{Minus-Plus Net (MP Net)}~\cite{xu2019recursive}: recursively removes the audio source that has the highest energy. \\
\noindent \textbf{Co-Separation}~\cite{gao2019co}: incorporates an object-level separation loss while training using the ``mix-and-separate'' framework. However, the visual conditioning is derived using only a single object in the scene.\\
\noindent \textbf{Sound of Motion (SofM)}~\cite{zhao2019sound}: integrates pixel-level motion trajectory and object/human appearances across video frames.\\
\noindent \textbf{Music Gesture (MG)}~\cite{gan2020music}: the most recent method on musical sound source separation, integrates appearance features from the scene along with human pose features. However this added requirement of human pose, limits its usability as a baseline to only the MUSIC dataset.\\

\subsection{Evaluation Metrics}
In order to quantify the performance of the different algorithms, we report the model performances in terms of the Signal-to-Distortion Ratio (SDR) [dB] \cite{Vincent2006BSSeval,raffel2014mir_eval}, where higher SDR indicates more faithful reproduction of the original signal. We also report two related measures, Signal-to-Interference Ratio (SIR) (which gives an indication of the amount of reduction of interference in the estimated signal) and  Signal-to-Artifact Ratio (SAR) (which gives an indication of how much artifacts were introduced), as they were reported in prior audio-visual separation works \cite{zhao2018sound,gao2019co}. 

\subsection{Implementation Details}
We implement our model in PyTorch \cite{pytorch}. Following prior works \cite{zhao2018sound,gao2019co}, we sub-sample the audio at 11 kHz, and compute the STFT of the audio using a Hann window of size 1022 and a hop length of 256. With input samples of length approximately 6 s, this yields a spectrogram of dimensions $512 \times 256$. The spectrogram is re-sampled according to a log-frequency scale to obtain a magnitude spectrogram of size $\Omega \times T$ with $\Omega=256, T=256$.
The detector for the musical instruments was trained on the 15 musical object categories of the Open Images Dataset \cite{kuznetsova2020open}. The FRCNN feature vectors $F$ are 2048 dimensional. We detect up to two principal objects per video and use a set of up to 20 context nodes for a principal object. Additionally, a random crop from the image is considered as another principal object and is considered as belonging to the ``background'' class. The IoU threshold is set to $\delta=0.1$, and 4 multi-head attention units are used in the graph attention network. The embedding dimension obtained from the graph pooling stage is set to $512$. The GRU used is unidirectional with one hidden layer of $512$ dimensions, and the visual representation vector thus has $d=512$ dimensions. The weights on the loss terms are set to $\lambda_1=1, \lambda_2=0.05, \lambda_3=1$. The model is trained using the ADAM optimizer~\cite{kingma2014adam} with a weight decay of $1\text{e-}{4}$, $\beta_1=0.9$, $\beta_2=0.999$. During training, the FRCNN model weights are frozen. An initial learning rate of $1\text{e-}{4}$ is used and is decreased by a factor of 0.1 after every 15,000 iterations. These hyper-parameters and those of the baseline models are chosen based on the performances over the respective validation sets of the two datasets. At test time, the visual graph corresponding to a video is paired with a mixed audio (obtained from one or multiple videos) and fed as input to the network, which iteratively separates the audio sources from the input audio signal. We then apply an inverse STFT transform to map the separated spectrogram to the time domain, for evaluation.

\begin{figure}[t]
    \centering
    \includegraphics[scale=0.3]{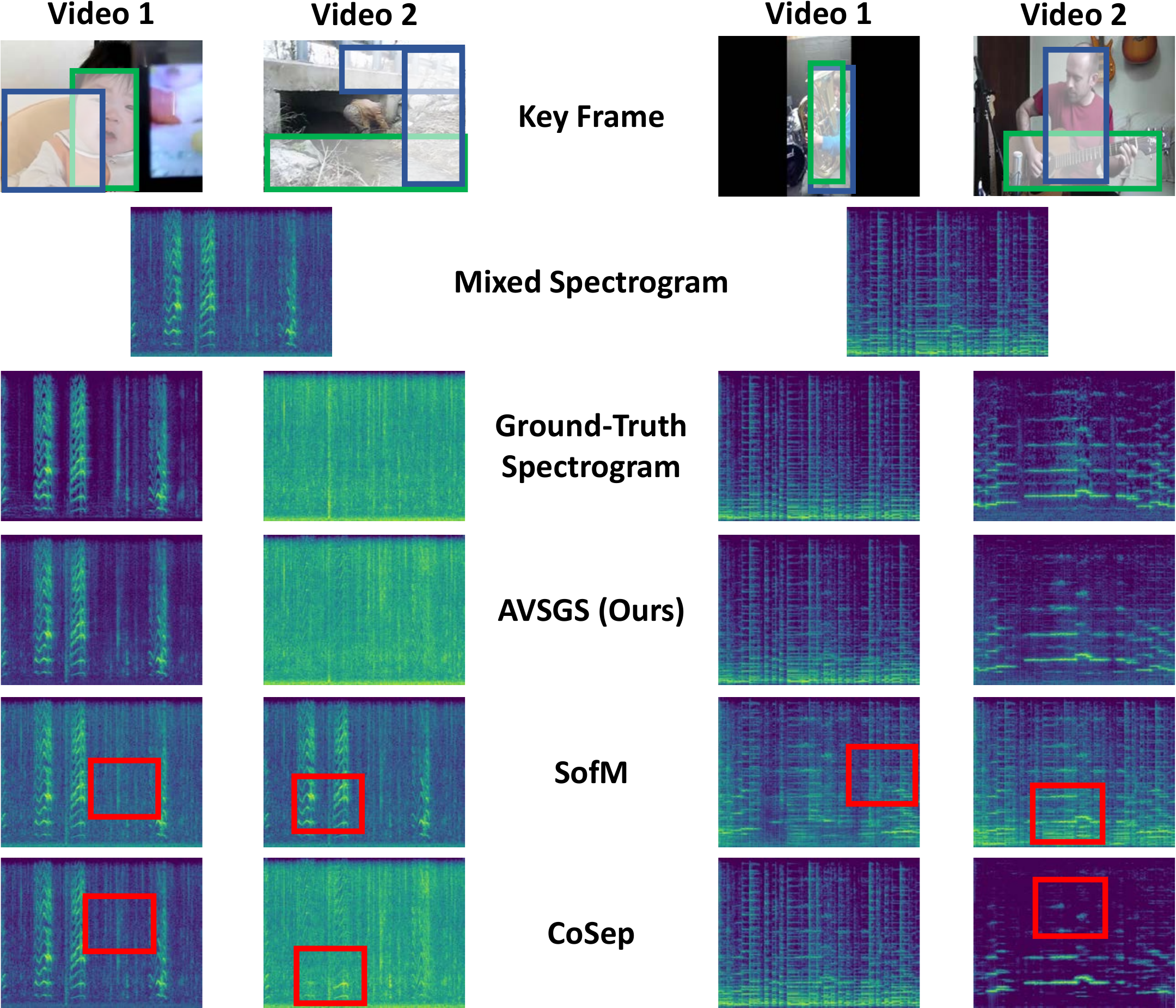} 
   \caption{ Qualitative separation results on ASIW (left) and MUSIC (right). Bounding boxes on frames show regions attended by AVSGS (green: principal object, blue: context nodes). Red boxes indicate regions of high differences between ground truth and predicted spectrograms.}
    \label{fig:model_perf}
\end{figure}
\begin{figure}[t]
    \centering
    \includegraphics[scale=0.2]{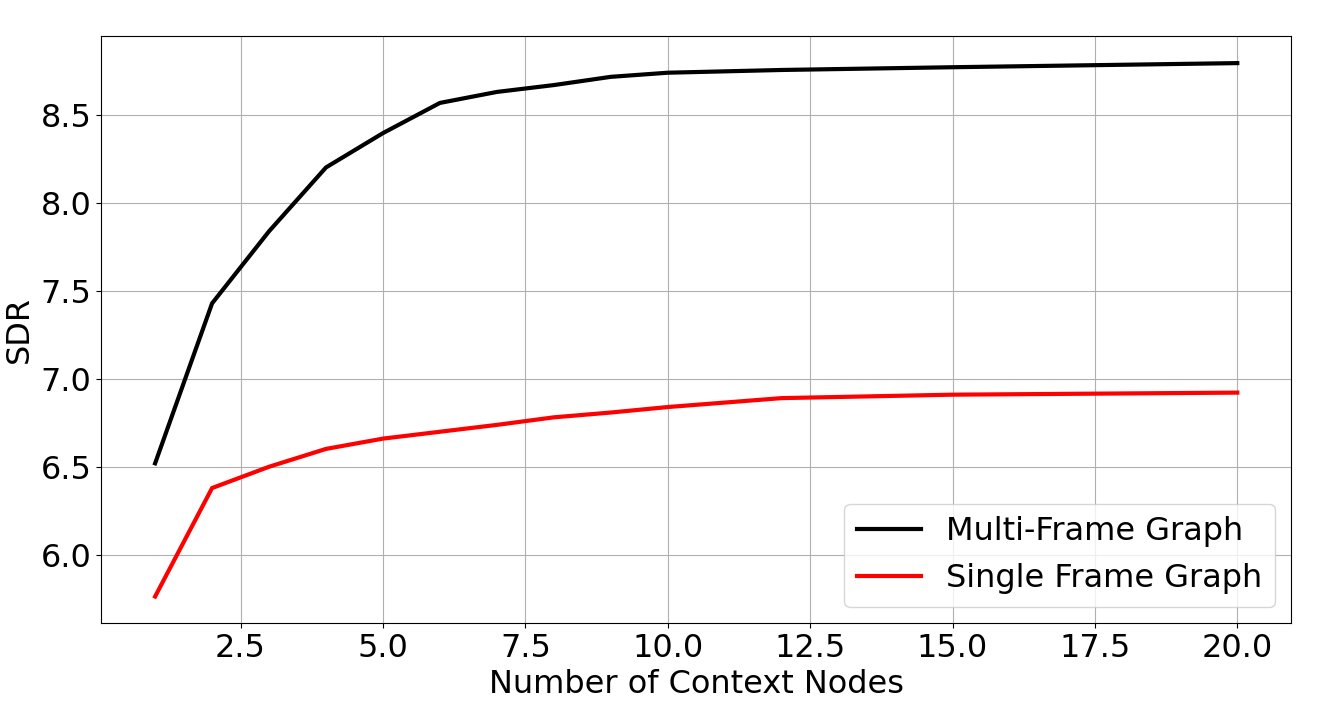} 
   \caption{ Performance plots of model variants with graphs constructed from multiple frames (black) and single frame (red).}
    \label{fig:perf_plot}
    \vspace*{-0.5cm}
\end{figure}

\subsection{Results}
We present the model performances on the MUSIC and ASIW datasets in Table~\ref{tab:perf_asiw_music}. From the results, we see that our proposed AVSGS model outperforms its closest competitor by large margins of around 1.3 dB on SDR and 1.6 dB on SIR on the MUSIC dataset, and around 1.2 dB on SDR and 2 dB on SIR on the ASIW dataset, which reflects substantial gains, given that these metrics are in log scale. We note that our much higher SIR did not come at the expense of a lower SAR, as is often the case, with the SAR in fact surpassing MG's~\cite{gan2020music} by 0.6 dB.
SofM is the non-MUSIC-specific baseline that comes closest to our model's performance, perhaps because it effectively combines motion and appearance, while the visual information of most other approaches is mainly appearance-based and holistic. In AVSGS, while motion is not explicitly encoded in the visual representation, the spatio-temporal nature of our graph $G$ implicitly embeds this key element. MG's competitive performance on MUSIC gives credence to the hypothesis that for good audio separation, besides the embedding of the principal object, appropriate visual context is necessary. In their setting, this is limited to human pose. However, when the set of context nodes is expanded, richer interactions can be captured, which proves beneficial to the model performance, as is seen to be the case for our model. Importantly, our approach for incorporating this context information generalizes well across both datasets, unlike MG. 

\noindent \textbf{Ablations and Additional Results}:
In Table~\ref{tab:asiw_ablation}, we report performances of several ablated variants of our model on the ASIW dataset. The second, third, and fourth rows showcase the model performance obtained by turning off one loss term at a time.  The results overwhelmingly point to the importance of the co-separation loss (row 3), without which the performance of the model drops significantly. We also tweaked the number of objects per video to 3 and observed very little change in model performance, as seen in row 5 of Table~\ref{tab:asiw_ablation}. Row 6 underscores the importance of having skip connection in the ASN network. In rows 7 and 8, we present results of ablating the different components of the scene graph. The results indicate that GATConv and EdgeConv are roughly equally salient. Finally, as seen in row 9, our model underperforms without the GRU.

\noindent Additionally, in Figure ~\ref{fig:perf_plot} we plot the performance of our AVSGS model with varying number of context nodes at test time, shown in black. This experiment is then repeated for a model where we build the graph from only a single frame. The performance plot of this variant is shown in red. The plots show a monotonically increasing trend underscoring the importance of constructing spatio-temporal graphs which capture the richness of the scene contexts.

\noindent \textbf{Qualitative Results}: In Figure~\ref{fig:model_perf}, we present example separation results on samples from the ASIW and MUSIC test sets, while contrasting the performance of our algorithm against two competitive baselines, Co-Separation and SofM. As is evident from the separated spectrograms, AVSGS is more effective in separating the sources than these baselines. Additionally, the figure also shows the regions attended to by AVSGS in order to induce the audio source separation. We find that AVSGS correctly chooses useful context regions/objects to attend to for both datasets. For more details, qualitative results, and user study, please see the supplementary materials.

\section{Conclusions}
We presented AVSGS, a novel algorithm that leverages the power of scene graphs to induce audio source separation. Our model leverages self-supervised techniques for training and does not require additional labelled training  data. We show that the added context information that the scene graphs introduce allows us to obtain state-of-the-art results on the existing MUSIC dataset and a challenging new dataset of ``in the wild'' videos called ASIW. In future work, we intend to explicitly incorporate motion into the scene graph to further boost model performance.

\vspace{4pt}
\noindent \textbf{Acknowledgements.} MC and NA would like to thank the support of the Office of Naval Research under grant N00014- 20-1-2444, and USDA National Institute of Food and Agriculture under grant 2020-67021-32799/1024178.

{\small
\bibliographystyle{ieee_fullname}
\bibliography{avsg-iccv21}

\begin{thebibliography}{10}\itemsep=-1pt

\bibitem{Afouras20b}
Triantafyllos Afouras, Andrew Owens, Joon~Son Chung, and Andrew Zisserman.
\newblock Self-supervised learning of audio-visual objects from video.
\newblock In {\em Proc. ECCV}, 2020.

\bibitem{arandjelovic2018objects}
Relja Arandjelovic and Andrew Zisserman.
\newblock Objects that sound.
\newblock In {\em Proc. ECCV}, pages 435--451, 2018.

\bibitem{chung2014empirical}
Junyoung Chung, Caglar Gulcehre, KyungHyun Cho, and Yoshua Bengio.
\newblock Empirical evaluation of gated recurrent neural networks on sequence
  modeling.
\newblock {\em arXiv preprint arXiv:1412.3555}, 2014.

\bibitem{comon2010handbook}
Pierre Comon and Christian Jutten.
\newblock {\em Handbook of Blind Source Separation: Independent component
  analysis and applications}.
\newblock Academic press, 2010.

\bibitem{ephrat2018looking}
Ariel Ephrat, Inbar Mosseri, Oran Lang, Tali Dekel, Kevin Wilson, Avinatan
  Hassidim, William~T Freeman, and Michael Rubinstein.
\newblock Looking to listen at the cocktail party: a speaker-independent
  audio-visual model for speech separation.
\newblock {\em ACM Trans. Graph. (TOG)}, 37(4):1--11, 2018.

\bibitem{gan2020music}
Chuang Gan, Deng Huang, Hang Zhao, Joshua~B Tenenbaum, and Antonio Torralba.
\newblock Music gesture for visual sound separation.
\newblock In {\em Proc. CVPR}, pages 10478--10487, 2020.

\bibitem{gannot2017consolidated}
Sharon Gannot, Emmanuel Vincent, Shmulik Markovich-Golan, and Alexey Ozerov.
\newblock A consolidated perspective on multimicrophone speech enhancement and
  source separation.
\newblock {\em IEEE/ACM Trans. Audio, Speech, Lang. Process.}, 25(4):692--730,
  2017.

\bibitem{gao2018learning}
Ruohan Gao, Rogerio Feris, and Kristen Grauman.
\newblock Learning to separate object sounds by watching unlabeled video.
\newblock In {\em Proc. ECCV}, pages 35--53, Sept. 2018.

\bibitem{gao20192}
Ruohan Gao and Kristen Grauman.
\newblock 2.5 d visual sound.
\newblock In {\em Proc. CVPR}, pages 324--333, 2019.

\bibitem{gao2019co}
Ruohan Gao and Kristen Grauman.
\newblock Co-separating sounds of visual objects.
\newblock In {\em Proc. ICCV}, pages 3879--3888, 2019.

\bibitem{gemmeke2017audio}
Jort~F Gemmeke, Daniel~PW Ellis, Dylan Freedman, Aren Jansen, Wade Lawrence,
  R~Channing Moore, Manoj Plakal, and Marvin Ritter.
\newblock Audio set: An ontology and human-labeled dataset for audio events.
\newblock In {\em Proc. ICASSP}, pages 776--780, Mar. 2017.

\bibitem{geng2020spatio}
Shijie Geng, Peng Gao, Chiori Hori, Jonathan~Le Roux, and Anoop Cherian.
\newblock Spatio-temporal scene graphs for video dialog.
\newblock In {\em Proc. AAAI}, 2021.

\bibitem{he2016deep}
Kaiming He, Xiangyu Zhang, Shaoqing Ren, and Jian Sun.
\newblock Deep residual learning for image recognition.
\newblock In {\em Proc. CVPR}, pages 770--778, 2016.

\bibitem{hershey1999audio}
John Hershey and Javier Movellan.
\newblock Audio vision: Using audio-visual synchrony to locate sounds.
\newblock In {\em Proc. NIPS}, pages 813--819, Dec. 1999.

\bibitem{Hershey2016ICASSP03}
John~R. Hershey, Zhuo Chen, Jonathan {Le Roux}, and Shinji Watanabe.
\newblock Deep clustering: Discriminative embeddings for segmentation and
  separation.
\newblock In {\em Proc. ICASSP}, Mar. 2016.

\bibitem{huang2014deep}
Po-Sen Huang, Minje Kim, Mark Hasegawa-Johnson, and Paris Smaragdis.
\newblock Deep learning for monaural speech separation.
\newblock In {\em Proc. ICASSP}, pages 1562--1566, May 2014.

\bibitem{Isik2016}
Yusuf Isik, Jonathan {Le Roux}, Zhuo Chen, Shinji Watanabe, and John~R.
  Hershey.
\newblock Single-channel multi-speaker separation using deep clustering.
\newblock In {\em Proc. Interspeech}, pages 545--549, Sept. 2016.

\bibitem{jansson2017singing}
Andreas Jansson, Eric Humphrey, Nicola Montecchio, Rachel Bittner, Aparna
  Kumar, and Tillman Weyde.
\newblock Singing voice separation with deep {U}-net convolutional networks.
\newblock In {\em Proc. ISMIR}, Oct. 2017.

\bibitem{ji2020action}
Jingwei Ji, Ranjay Krishna, Li Fei-Fei, and Juan~Carlos Niebles.
\newblock Action genome: Actions as compositions of spatio-temporal scene
  graphs.
\newblock In {\em Proc. CVPR}, pages 10236--10247, 2020.

\bibitem{johnson2015image}
Justin Johnson, Ranjay Krishna, Michael Stark, Li-Jia Li, David Shamma, Michael
  Bernstein, and Li Fei-Fei.
\newblock Image retrieval using scene graphs.
\newblock In {\em Proc. CVPR}, pages 3668--3678, 2015.

\bibitem{kidron2005pixels}
Einat Kidron, Yoav~Y Schechner, and Michael Elad.
\newblock Pixels that sound.
\newblock In {\em Proc. CVPR}, volume~1, pages 88--95. IEEE, 2005.

\bibitem{kim2019audiocaps}
Chris~Dongjoo Kim, Byeongchang Kim, Hyunmin Lee, and Gunhee Kim.
\newblock Audiocaps: Generating captions for audios in the wild.
\newblock In {\em Proc. NAACL HLT}, pages 119--132, 2019.

\bibitem{kingma2014adam}
Diederik~P Kingma and Jimmy Ba.
\newblock Adam: A method for stochastic optimization.
\newblock In {\em Proc. ICLR}, 2014.

\bibitem{krasin2017openimages}
Ivan Krasin, Tom Duerig, Neil Alldrin, Vittorio Ferrari, Sami Abu-El-Haija,
  Alina Kuznetsova, Hassan Rom, Jasper Uijlings, Stefan Popov, Andreas Veit,
  et~al.
\newblock Openimages: A public dataset for large-scale multi-label and
  multi-class image classification.
\newblock {\em Dataset available from https://github. com/openimages}, 2(3):18,
  2017.

\bibitem{krishna2017visual}
Ranjay Krishna, Yuke Zhu, Oliver Groth, Justin Johnson, Kenji Hata, Joshua
  Kravitz, Stephanie Chen, Yannis Kalantidis, Li-Jia Li, David~A Shamma, et~al.
\newblock Visual {G}enome: Connecting language and vision using crowdsourced
  dense image annotations.
\newblock {\em International Journal of Computer Vision}, 123(1):32--73, 2017.

\bibitem{kuznetsova2020open}
Alina Kuznetsova, Hassan Rom, Neil Alldrin, Jasper Uijlings, Ivan Krasin, Jordi
  Pont-Tuset, Shahab Kamali, Stefan Popov, Matteo Malloci, Alexander
  Kolesnikov, et~al.
\newblock The {O}pen {I}mages dataset v4.
\newblock {\em Int. J. Comput. Vis.}, pages 1--26, 2020.

\bibitem{lee2019self}
Junhyun Lee, Inyeop Lee, and Jaewoo Kang.
\newblock Self-attention graph pooling.
\newblock In {\em Proc. ICML}, pages 3734--3743, June 2019.

\bibitem{li2008modeling}
Xiaowei Li, Changchang Wu, Christopher Zach, Svetlana Lazebnik, and Jan-Michael
  Frahm.
\newblock Modeling and recognition of landmark image collections using iconic
  scene graphs.
\newblock In {\em Proc. ECCV}, pages 427--440. Springer, 2008.

\bibitem{li2009optimality}
Yipeng Li and DeLiang Wang.
\newblock On the optimality of ideal binary time-frequency masks.
\newblock {\em Speech Communication}, 51(3):230--239, 2009.

\bibitem{liu2019divide}
Yuzhou Liu and DeLiang Wang.
\newblock Divide and conquer: A deep {CASA} approach to talker-independent
  monaural speaker separation.
\newblock {\em IEEE/ACM Trans. Audio, Speech, Lang. Process.},
  27(12):2092--2102, 2019.

\bibitem{loizou2013speech}
Philipos~C Loizou.
\newblock {\em Speech enhancement: theory and practice}.
\newblock CRC press, 2013.

\bibitem{meseguer2019conditioned}
Gabriel Meseguer-Brocal and Geoffroy Peeters.
\newblock Conditioned-{U}-{N}et: Introducing a control mechanism in the
  {U}-{N}et for multiple source separations.
\newblock {\em arXiv preprint arXiv:1907.01277}, 2019.

\bibitem{michelsanti2020overview}
Daniel Michelsanti, Zheng-Hua Tan, Shi-Xiong Zhang, Yong Xu, Meng Yu, Dong Yu,
  and Jesper Jensen.
\newblock An overview of deep-learning-based audio-visual speech enhancement
  and separation.
\newblock {\em arXiv preprint arXiv:2008.09586}, 2020.

\bibitem{morgado2018self}
Pedro Morgado, Nuno Vasconcelos, Timothy Langlois, and Oliver Wang.
\newblock Self-supervised generation of spatial audio for 360° video.
\newblock In {\em Proc. NeurIPS}, pages 360--370, 2018.

\bibitem{owens2018audio}
Andrew Owens and Alexei~A Efros.
\newblock Audio-visual scene analysis with self-supervised multisensory
  features.
\newblock In {\em Proc. ECCV}, pages 631--648, 2018.

\bibitem{owens2016visually}
Andrew Owens, Phillip Isola, Josh McDermott, Antonio Torralba, Edward~H
  Adelson, and William~T Freeman.
\newblock Visually indicated sounds.
\newblock In {\em Proc. CVPR}, pages 2405--2413, 2016.

\bibitem{pytorch}
Adam Paszke, Sam Gross, Francisco Massa, Adam Lerer, James Bradbury, Gregory
  Chanan, Trevor Killeen, Zeming Lin, Natalia Gimelshein, Luca Antiga, Alban
  Desmaison, Andreas Kopf, Edward Yang, Zachary DeVito, Martin Raison, Alykhan
  Tejani, Sasank Chilamkurthy, Benoit Steiner, Lu Fang, Junjie Bai, and Soumith
  Chintala.
\newblock {PyTorch}: An imperative style, high-performance deep learning
  library.
\newblock In {\em Proc. NeurIPS}, pages 8024--8035, Dec. 2019.

\bibitem{pishdadian2020finding}
Fatemeh Pishdadian, Gordon Wichern, and Jonathan Le~Roux.
\newblock Finding strength in weakness: Learning to separate sounds with weak
  supervision.
\newblock {\em IEEE/ACM Trans. Audio, Speech, Lang. Process.}, 28:2386--2399,
  2020.

\bibitem{raffel2014mir_eval}
Colin Raffel, Brian McFee, Eric~J Humphrey, Justin Salamon, Oriol Nieto, Dawen
  Liang, Daniel~PW Ellis, and C~Colin Raffel.
\newblock mir\_eval: A transparent implementation of common mir metrics.
\newblock In {\em Proc. ISMIR}, 2014.

\bibitem{ren2016faster}
Shaoqing Ren, Kaiming He, Ross Girshick, and Jian Sun.
\newblock Faster r-cnn: towards real-time object detection with region proposal
  networks.
\newblock {\em IEEE Trans. Pattern Anal. Mach. Intell.}, 39(6):1137--1149,
  2016.

\bibitem{ronneberger2015u}
Olaf Ronneberger, Philipp Fischer, and Thomas Brox.
\newblock U-net: Convolutional networks for biomedical image segmentation.
\newblock In {\em Proc. MICCAI}, pages 234--241. Springer, 2015.

\bibitem{senocak2018learning}
Arda Senocak, Tae-Hyun Oh, Junsik Kim, Ming-Hsuan Yang, and In~So Kweon.
\newblock Learning to localize sound source in visual scenes.
\newblock In {\em Proc. CVPR}, pages 4358--4366, 2018.

\bibitem{slizovskaia2019end}
Olga Slizovskaia, Leo Kim, Gloria Haro, and Emilia Gomez.
\newblock End-to-end sound source separation conditioned on instrument labels.
\newblock In {\em Proc. ICASSP}, pages 306--310, May 2019.

\bibitem{smaragdis2014static}
Paris Smaragdis, Cedric Fevotte, Gautham~J Mysore, Nasser Mohammadiha, and
  Matthew Hoffman.
\newblock Static and dynamic source separation using nonnegative
  factorizations: A unified view.
\newblock {\em IEEE Signal Process. Mag.}, 31(3):66--75, 2014.

\bibitem{tzinis2020into}
Efthymios Tzinis, Scott Wisdom, Aren Jansen, Shawn Hershey, Tal Remez,
  Daniel~PW Ellis, and John~R Hershey.
\newblock Into the wild with {AudioScope}: Unsupervised audio-visual separation
  of on-screen sounds.
\newblock In {\em Proc. ICLR}, 2021.

\bibitem{velivckovic2017graph}
Petar Veli{\v{c}}kovi{\'c}, Guillem Cucurull, Arantxa Casanova, Adriana Romero,
  Pietro Lio, and Yoshua Bengio.
\newblock Graph attention networks.
\newblock In {\em Proc. ICLR}, Apr. 2018.

\bibitem{Vincent2006BSSeval}
Emmanuel Vincent, R\'{e}mi Gribonval, and C\'{e}dric F\'{e}votte.
\newblock Performance measurement in blind audio source separation.
\newblock {\em IEEE Trans. Audio, Speech, Lang. Process.}, 14(4):1462--1469,
  July 2006.

\bibitem{vincent2018audio}
Emmanuel Vincent, Tuomas Virtanen, and Sharon Gannot.
\newblock {\em Audio source separation and speech enhancement}.
\newblock John Wiley \& Sons, 2018.

\bibitem{wang2006computational}
DeLiang Wang and Guy~J Brown.
\newblock {\em Computational auditory scene analysis: Principles, algorithms,
  and applications}.
\newblock Wiley-IEEE press, 2006.

\bibitem{wang2018supervised}
DeLiang Wang and Jitong Chen.
\newblock Supervised speech separation based on deep learning: An overview.
\newblock {\em IEEE/ACM Trans. Audio, Speech, Lang. Process.},
  26(10):1702--1726, 2018.

\bibitem{wang2019dynamic}
Yue Wang, Yongbin Sun, Ziwei Liu, Sanjay~E Sarma, Michael~M Bronstein, and
  Justin~M Solomon.
\newblock Dynamic graph {CNN} for learning on point clouds.
\newblock {\em ACM Trans. Graph. (TOG)}, 38(5):1--12, 2019.

\bibitem{Weninger2014GlobalSIP12}
Felix Weninger, Jonathan {Le Roux}, John~R. Hershey, and Bj{\"o}rn Schuller.
\newblock Discriminatively trained recurrent neural networks for single-channel
  speech separation.
\newblock In {\em Proc. GlobalSIP}, Dec. 2014.

\bibitem{weninger2014nmf}
Felix Weninger, Jonathan~Le Roux, John~R Hershey, and Shinji Watanabe.
\newblock Discriminative {NMF} and its application to single-channel source
  separation.
\newblock In {\em Proc. Interspeech}, Sept. 2014.

\bibitem{wisdom2020unsupervised}
Scott Wisdom, Efthymios Tzinis, Hakan Erdogan, Ron~J Weiss, Kevin Wilson, and
  John~R Hershey.
\newblock Unsupervised sound separation using mixtures of mixtures.
\newblock In {\em Proc. NeurIPS}, Dec. 2020.

\bibitem{xia2017using}
Shasha Xia, Hao Li, and Xueliang Zhang.
\newblock Using optimal ratio mask as training target for supervised speech
  separation.
\newblock In {\em Proc. APSIPA ASC}, pages 163--166, 2017.

\bibitem{xu2019recursive}
Xudong Xu, Bo Dai, and Dahua Lin.
\newblock Recursive visual sound separation using minus-plus net.
\newblock In {\em Proc. ICCV}, pages 882--891, Oct. 2019.

\bibitem{xu2013experimental}
Yong Xu, Jun Du, Li-Rong Dai, and Chin-Hui Lee.
\newblock An experimental study on speech enhancement based on deep neural
  networks.
\newblock {\em IEEE Signal Process. Lett.}, 21(1):65--68, 2013.

\bibitem{yu2017permutation}
Dong Yu, Morten Kolb{\ae}k, Zheng-Hua Tan, and Jesper Jensen.
\newblock Permutation invariant training of deep models for speaker-independent
  multi-talker speech separation.
\newblock In {\em Proc. ICASSP}, pages 241--245, Mar. 2017.

\bibitem{zhao2019sound}
Hang Zhao, Chuang Gan, Wei-Chiu Ma, and Antonio Torralba.
\newblock The sound of motions.
\newblock In {\em Proc. ICCV}, pages 1735--1744, 2019.

\bibitem{zhao2018sound}
Hang Zhao, Chuang Gan, Andrew Rouditchenko, Carl Vondrick, Josh McDermott, and
  Antonio Torralba.
\newblock The sound of pixels.
\newblock In {\em Proc. ECCV}, pages 570--586, 2018.

\bibitem{zhou2018visual}
Yipin Zhou, Zhaowen Wang, Chen Fang, Trung Bui, and Tamara~L Berg.
\newblock Visual to sound: Generating natural sound for videos in the wild.
\newblock In {\em Proc. CVPR}, pages 3550--3558, 2018.

\end{thebibliography}
}

\appendix
\section{ASIW Dataset Details}
\label{sec:asiw_dataset}

\begin{table*}[t]
 \centering
 \caption{Number of videos for each of the principal object categories of ASIW dataset. }\label{tab:asiw_principstats}
 {\setlength{\tabcolsep}{4pt}
 \begin{tabular}{cccccccccccccc}
\toprule
  {Baby}	& {Bell} &	{Birds} &	{Camera} &	{Clock} &	{Dogs} &	{Toilet} &	{Horse} &	{Man} &	{Sheep} &	{Telephone} &	{Trains} &	{Vehicle} &	{Water} \\ 
  \midrule
    1616	& 151 &	2887 &	913 &	658 &	1407 &	838 &	385 &	6210 &	710 &	222 &	141 &	779 &	378 \\ 
    \bottomrule
\end{tabular}
}
\vspace{-0.3cm}
\end{table*}

 Most prior approaches in visually-guided sound source separation report performances solely in the setting of separating the sounds of musical instruments ~\cite{gao2019co,zhao2018sound,zhao2019sound,gan2020music}. However, musical instruments often have very characteristic sounds and thereby the range of variability within a particular instrument category is limited. Moreover, the videos featured in these datasets are often recorded professionally in rather controlled environments, such as an auditorium. Such videos however, may not capture the variety of sounds that we come across in daily-life settings. In order to fill this void, this work introduces the \textit{Audio Separation in the Wild (ASIW)} dataset. 
 
 ASIW is adapted from the recently introduced large-scale AudioCaps dataset \cite{kim2019audiocaps}, which contains 49,838 training, 495 validation, and 975 test videos crawled from the AudioSet dataset \cite{gemmeke2017audio}, each of which is about 10s long. These videos have been carefully captioned manually (by English-speaking Amazon Mechanical Turkers -- AMTs). In comparison to other video captioning datasets (such as MSVD or MSRVTT), AudioCaps captions are particularly focused on describing auditory events in the video; which motivated us to consider this dataset for the task of visually-guided sound source separation.

 \begin{figure}[t]
    \centering
    \includegraphics[scale=0.2]{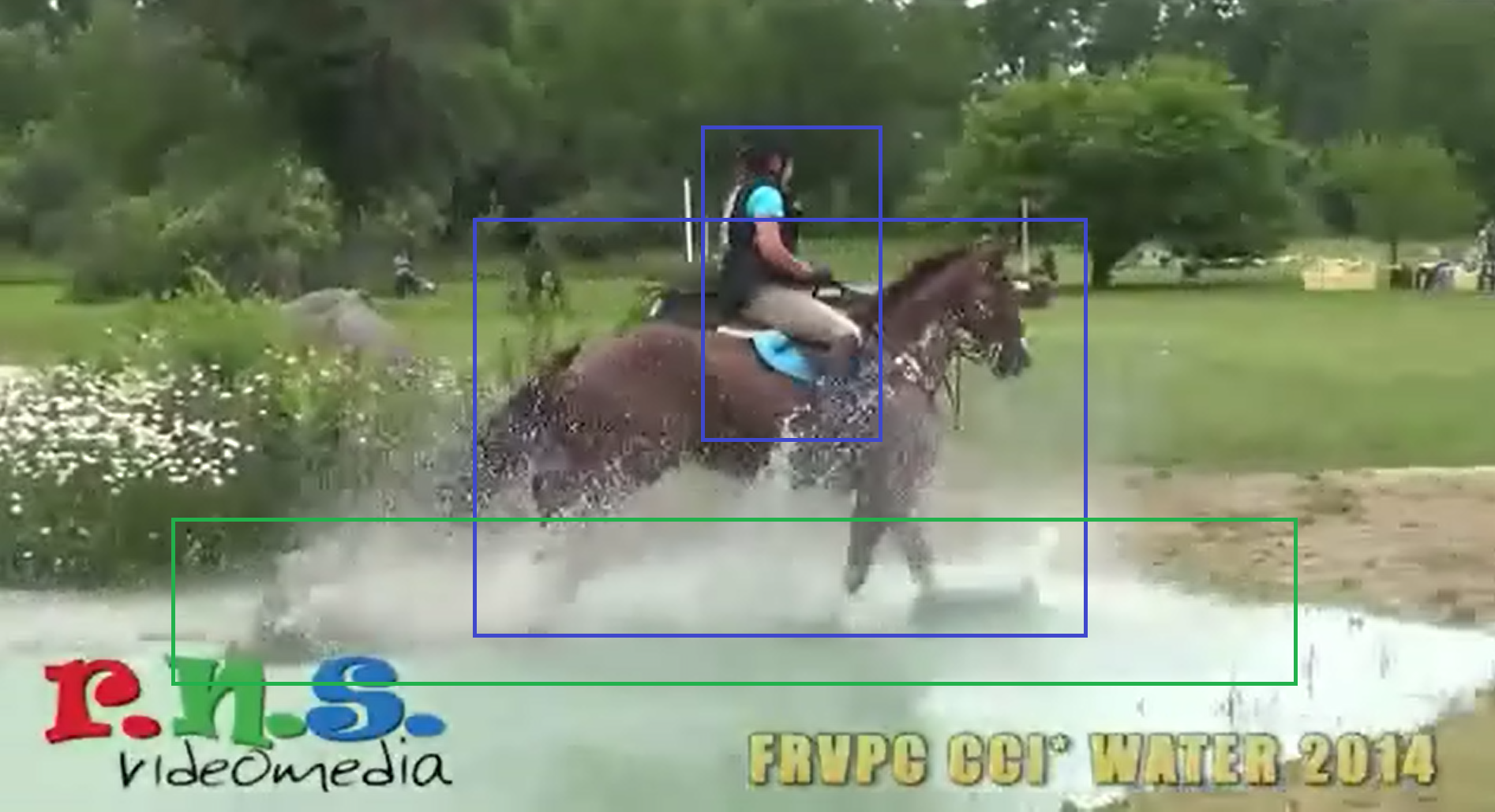} 
   \caption{ A sample frame from a video in the ASIW dataset, showing the principal object (water, highlighted by a green box) and a set of interacting objects (\emph{horse} with a \emph{rider}, highlighted by a blue box).}
    \label{fig:asiw_sample}
    \vspace*{-0.5cm}
\end{figure}
 
To adapt AudioCaps for our task, we manually construct a dictionary of 306 frequently occurring \textit{auditory words} from the captions, such as: \textit{splashing}, \textit{flushing}, \textit{eruptions}, or \textit{giggling}. Another factor we considered in order to select this dictionary is the grounding that the words have in the video; which we call the \emph{pricipal objects} in the main paper. The words in the dictionary are selected such that they have a corresponding principal object in the video generating the respective sound. The set of principal objects that we finally selected from AudioCaps consisted of 14 classes, namely: \emph{baby}, \emph{bell}, \emph{birds}, \emph{camera}, \emph{clock}, \emph{dogs}, \emph{toilet}, \emph{horse}, \emph{man/woman}, \emph{sheep/goat}, \emph{telephone}, \emph{trains}, \emph{vehicle/car/truck}, \emph{water}, and an additional \emph{background} class, which encompasses words that usually do not  consistently ground to a visible principal object in the video. For instance, \emph{brushing} could ground to a person brushing his/her teeth with a toothbrush or could also map to a painter putting his/her strokes on a canvas. We construct the \emph{principal object list} from the Visual Genome \cite{krishna2017visual} classes. The number of videos in each of these classes is shown in Table~\ref{tab:asiw_principstats}. In Figure~\ref{fig:asiw_sample}, we show a sample frame from a video in this dataset, highlighting the principal object (in green) -  in this case \emph{water} interacting with  another object (in blue), viz. \emph{a horse with a jockey}, to produce the auditory word \textit{splashing}.
 
In Section~\ref{sec:listofwords}, we list the full set of auditory words (in \textbf{bold-face} font), indicating alongside which principal object it is grounded to as well as its frequency in the captions associated with the dataset. While constructing the dataset, all principal object classes which consistently exhibit the same sound are treated as the same class and are indicated in the above list in the same row, separated by a forward-slash ('/'). For instance, although the class ``clock'' is different from the class ``clock tower'', visually, but since a possible sound emitted by both may be characterized as ``donging'',  we treat them as equivalent principal objects. We intend to make this dataset publicly available for researchers in the community, upon the acceptance of this work.

\section{Network Architecture Details}
\label{sec:net_arch}
Our model, the \emph{\fullname}~(\name) has several components. Below, we list the key details of each of the components.

\subsection{Feature Extractor}
Our model commences with extracting features, corresponding to bounding boxes in the scene. In order to do so, we use a Faster R-CNN model~\cite{ren2016faster}, with a ResNet-101~\cite{he2016deep} backbone pre-trained on the Visual Genome Dataset~\cite{krishna2017visual}. In order to obtain instrument features for the MUSIC dataset another detector~\cite{gao2019co} is trained on the the OpenImages dataset~\cite{krasin2017openimages}. The former gives 2048-dimensional vectors, while the latter gives 512-dimensional vectors. In order to maintain consistency of feature dimensions across objects, we further encode the 2048-dimensional vectors into 512-dimensions through a 2-layer Multi-layer perceptron with Leaky ReLU activations (negative slope=0.2)

\subsection{Graph Attention Network}
Post the object detection and feature extraction, the scene-graph is constructed following the method laid out in the \textit{Proposed Method} section of the paper. The scene graph is then processed by a \textit{Graph Attention Network}, which has a cascade of the following three components:

\noindent \textbf{Graph Attention Network Convolution}: The Graph Attention Network Convolution (GATConv) ~\cite{velivckovic2017graph} updates the node features of the graph based on the edge adjacency information by applying multi-head graph message-passing. We use 4 heads in the network and the dimension of the output feature of this network is 512.

\noindent \textbf{Edge Convolution}: Next, we employ Edge Convolutions~\cite{wang2019dynamic} to capture pair-wise interactions, which take in a concatenated vector of 2 objects ($512 \times 2 = 1024$) and generates a 512-dimensional vector.

\noindent \textbf{Pooling Layers}: The final step of the \textit{Graph Attention Network} consists of pooling these feature vectors~\cite{lee2019self} to obtain a single vector. We concatenate the embeddings obtained by Global Max and Average Pool to obtain this.

\subsection{Recurrent Network}
Our Recurrent Network is instantiated via a \textit{Gated Recurrent Unit} (GRU)~\cite{chung2014empirical}, whose input space and feature dimensions are 512-dimensional.

\subsection{Audio Separator Network}
A key component of our model is the audio separator network that takes as input a mixed audio track and produces a separated sound source as output, conditioned on a visual feature. The network roughly follows a U-Net~\cite{ronneberger2015u} style architecture, with the visual feature being concatenated into the network at the bottleneck layer. The network has 7 convolution and 7 up-convolution layers, each with $4 \times 4$ filter dimensions and LeakyRELU activations with negative slope of 0.2. Additionally, there are skip connections between a pair of layers in the encoder and the decoder, with matching spatial resolution of their feature maps. The bottleneck layer has $2 \times 2 \times 512$ dimension and thus the visual feature vector obtained from the pre-processing above is tiled $2 \times 2$ times and then concatenated into the network at the bottleneck layer, along the channel dimension.

\section{Qualitative Results}
\label{sec:qual_results}

In this section, we present separated spectrogram visualizations obtained by our method versus competing baselines on both datasets, for a qualitative assessment by the reader. To this end, we show spectrogram separations for audio obtained from a mix of two different videos as well as separations on videos which have a mixture of multiple sound sources. 

\subsection{Qualitative Visualizations}

\begin{figure}[t]
    \centering
    \includegraphics[width=0.99\columnwidth,trim={0cm 0cm 20cm 0cm},clip=True]{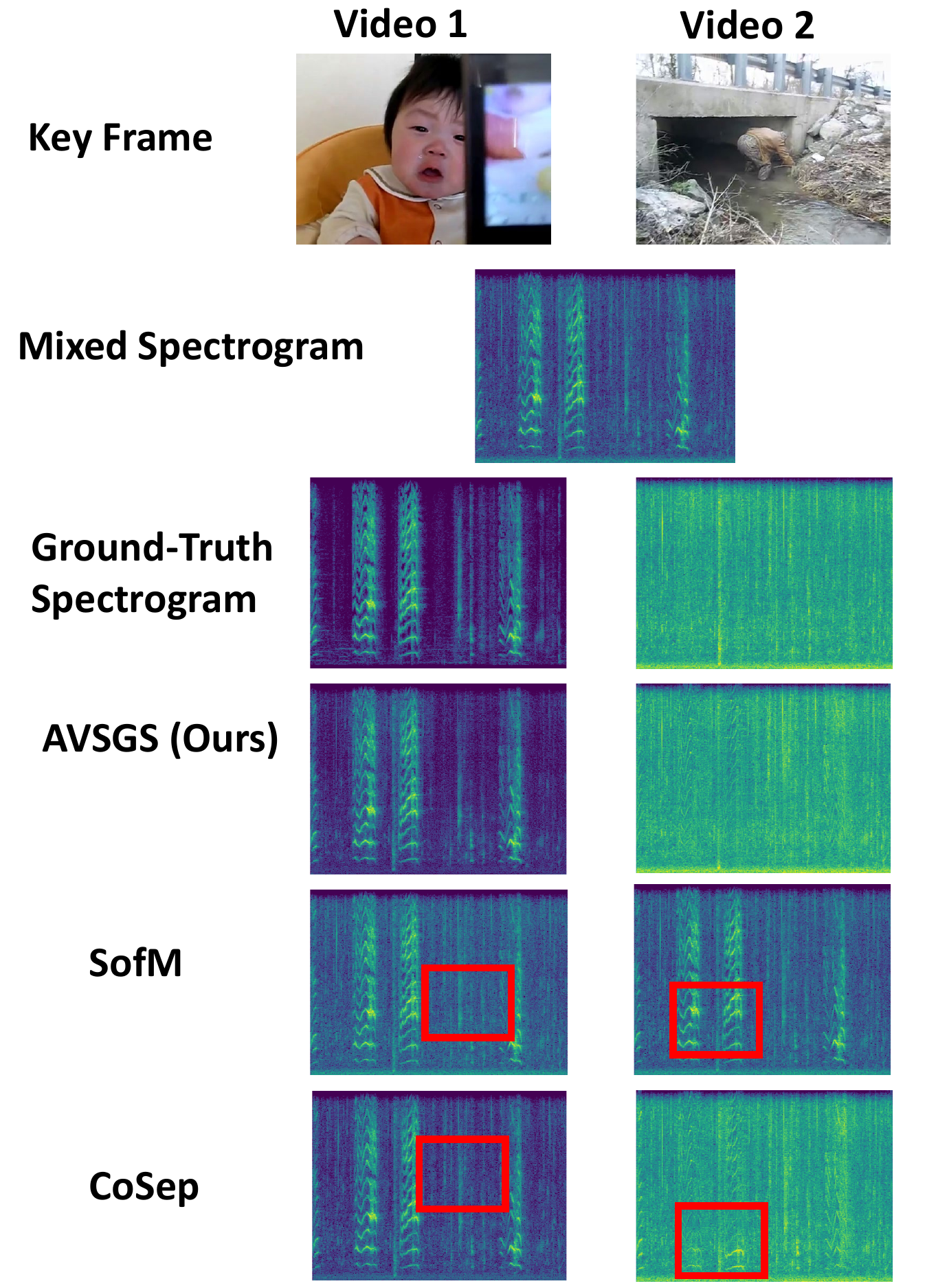} 
   \caption{ Qualitative separation results on a mixture of two ASIW videos. Sample key frames for both videos are shown. The spectrogram of the mixed audio is plotted as well. Also shown are the separated spectrograms obtained by different methods. Red boxes indicate regions of high differences between ground truth and predicted spectrograms.}
    \label{fig:asiw_perf_pair1}
\end{figure}

\begin{figure}[t]
    \centering
    \includegraphics[width=0.99\columnwidth,trim={0cm 0cm 20cm 0cm},clip=True]{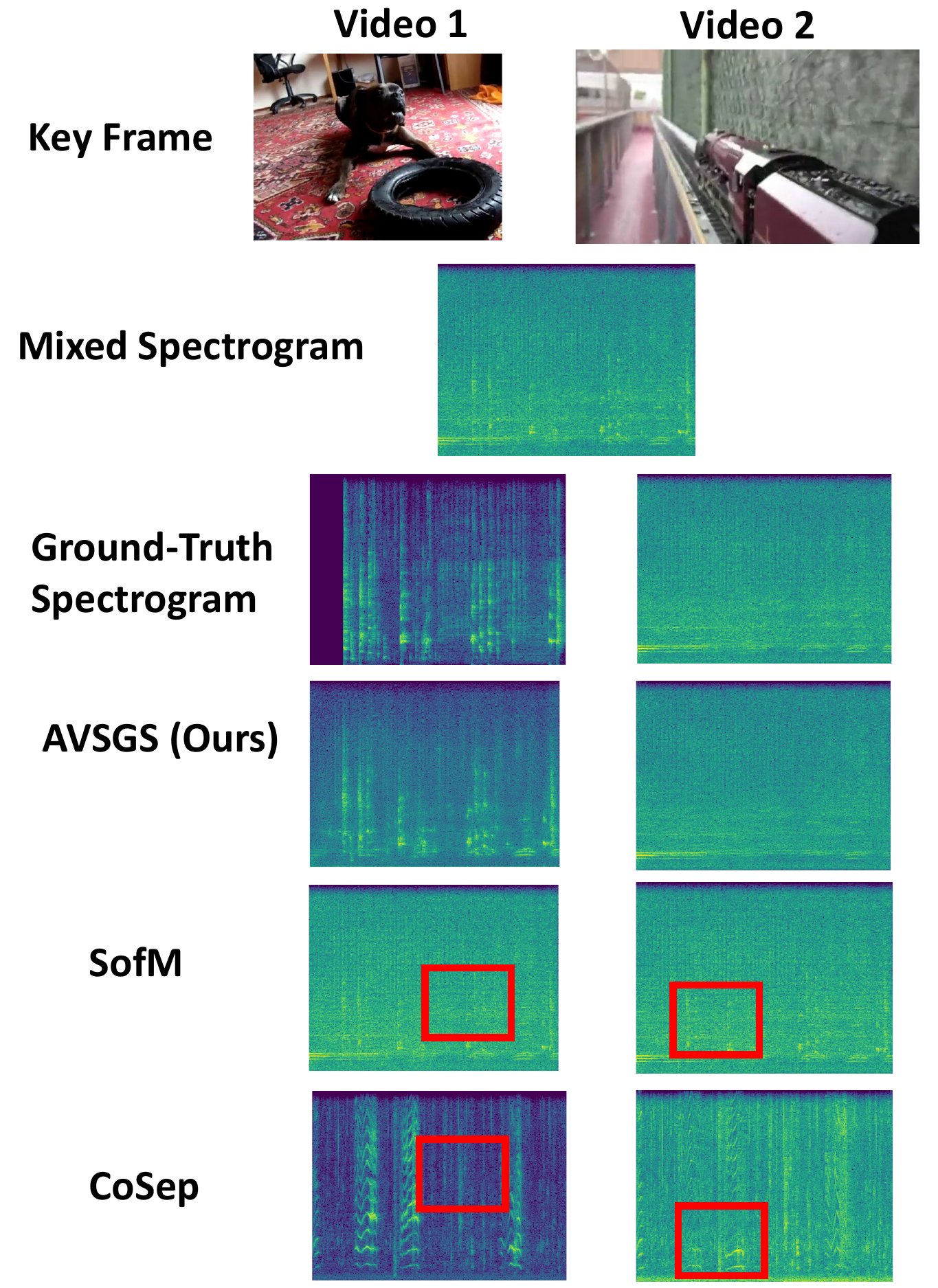} 
   \caption{ Qualitative separation results on a mixture of two ASIW videos. Sample key frames for both videos are shown. The spectrogram of the mixed audio is plotted as well. Also shown are the separated spectrograms obtained by different methods. Red boxes indicate regions of high differences between ground truth and predicted spectrograms.}
    \label{fig:asiw_perf_pair2}
\end{figure}

\begin{figure}[t]
    \centering
    \includegraphics[width=0.99\columnwidth,trim={0cm 0cm 20cm 0cm},clip=True]{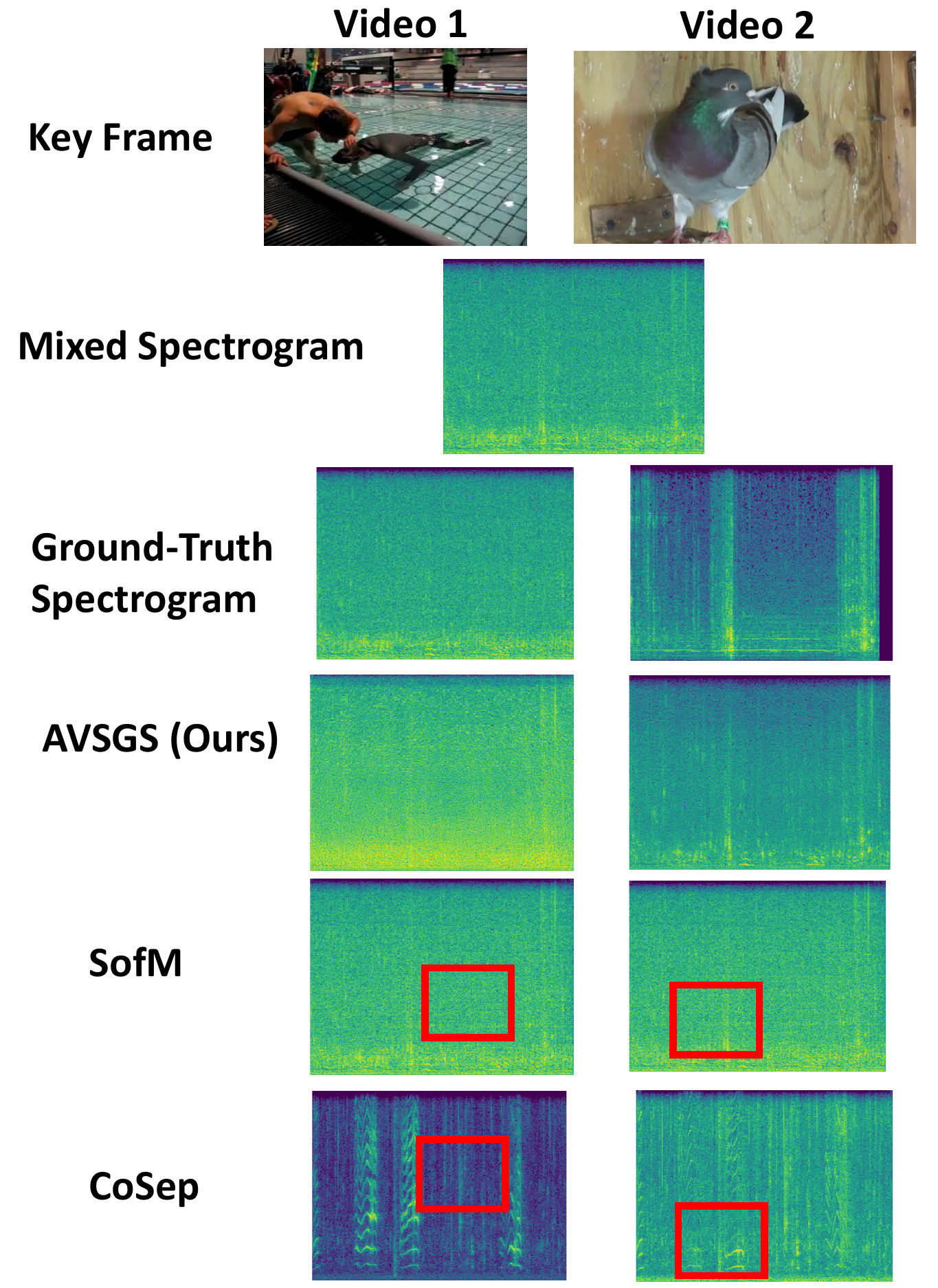} 
   \caption{ Qualitative separation results on a mixture of two ASIW videos. Sample key frames for both videos are shown. The spectrogram of the mixed audio is plotted as well. Also shown are the separated spectrograms obtained by different methods. Red boxes indicate regions of high differences between ground truth and predicted spectrograms.}
    \label{fig:asiw_perf_pair3}
\end{figure}

\begin{figure}[t]
    \centering
    \includegraphics[width=0.99\columnwidth,trim={0cm 0cm 20cm 0cm},clip=True]{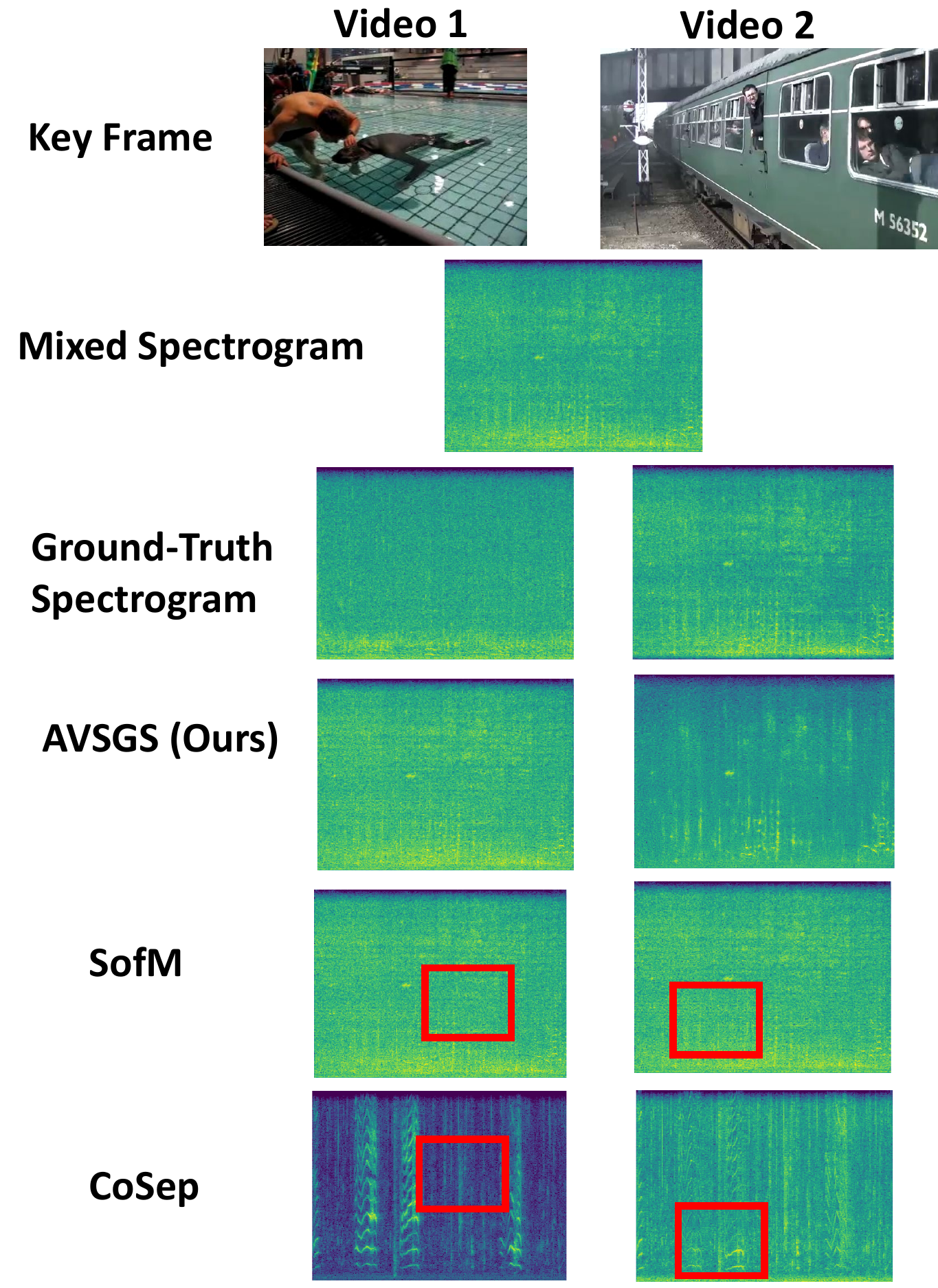} 
   \caption{ Qualitative separation results on a mixture of two ASIW videos. Sample key frames for both videos are shown. The spectrogram of the mixed audio is plotted as well. Also shown are the separated spectrograms obtained by different methods. Red boxes indicate regions of high differences between ground truth and predicted spectrograms.}
    \label{fig:asiw_perf_pair4}
\end{figure}

\begin{figure}[t]
    \centering
    \includegraphics[width=0.99\columnwidth,trim={0cm 0cm 20cm 0cm},clip=True]{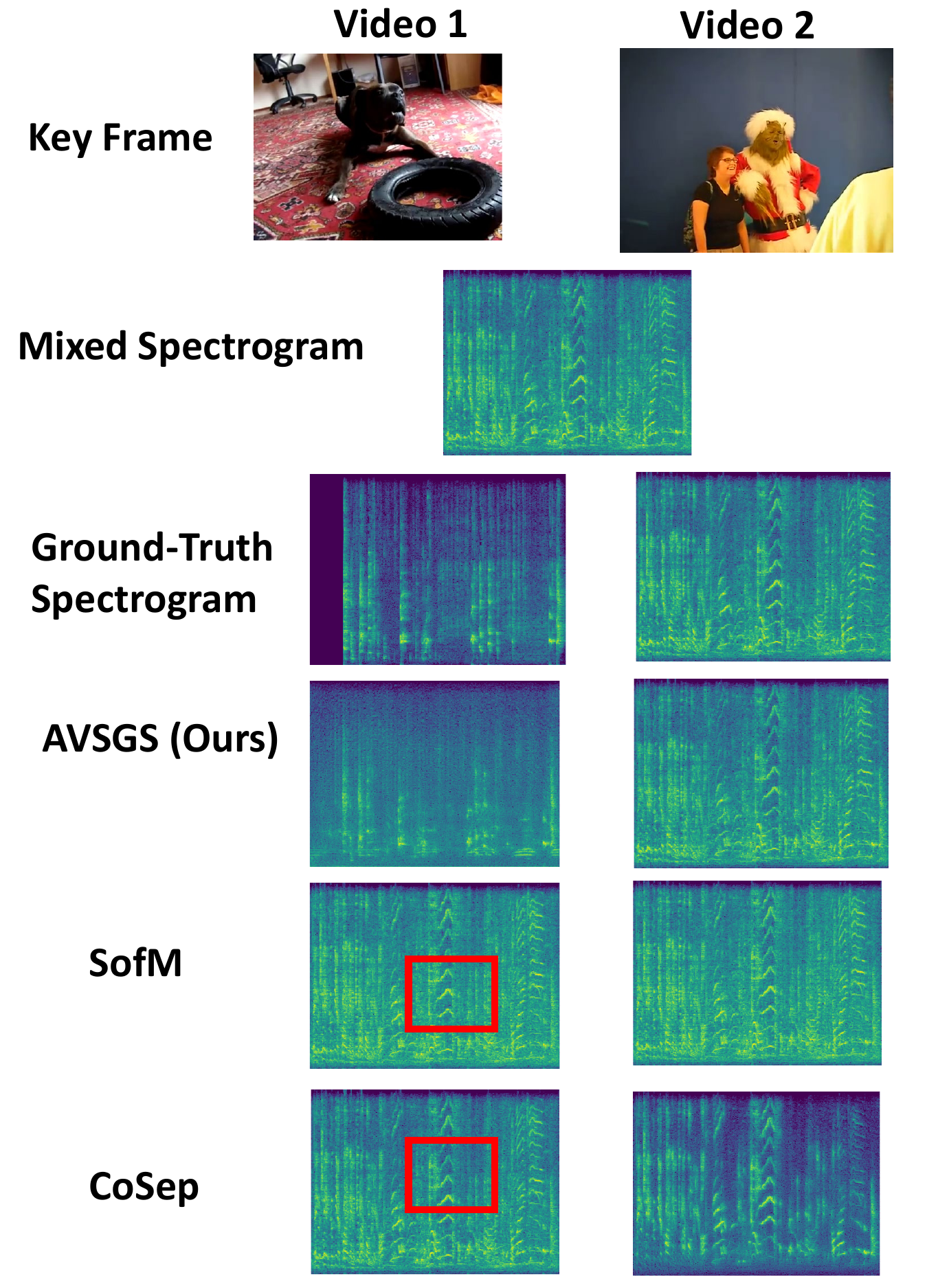} 
   \caption{ Qualitative separation results on a mixture of two ASIW videos. Sample key frames for both videos are shown. The spectrogram of the mixed audio is plotted as well. Also shown are the separated spectrograms obtained by different methods. Red boxes indicate regions of high differences between ground truth and predicted spectrograms.}
    \label{fig:asiw_perf_pair5}
\end{figure}

 \begin{table}[t]
 \centering
 \caption{Human preference score on samples from our method vs. Zhao \etal~\cite{zhao2019sound}}\label{tab:human_eval}
 \begin{tabular}{lc}
 \toprule
 \textbf{Datasets} & \textbf{ Prefer ours} \\ 
 \midrule
 ASIW  - Ours vs.~\cite{zhao2019sound}  & \textbf{92\%} \\ 
 MUSIC - Ours vs.~\cite{zhao2019sound} & \textbf{83\%} \\ 
 \bottomrule
 \end{tabular}
 \vspace{-0.2cm}
 \end{table}

\begin{figure}[t]
    \centering
    \includegraphics[width=0.99\columnwidth,trim={0cm 6cm 20cm 0cm},clip=True]{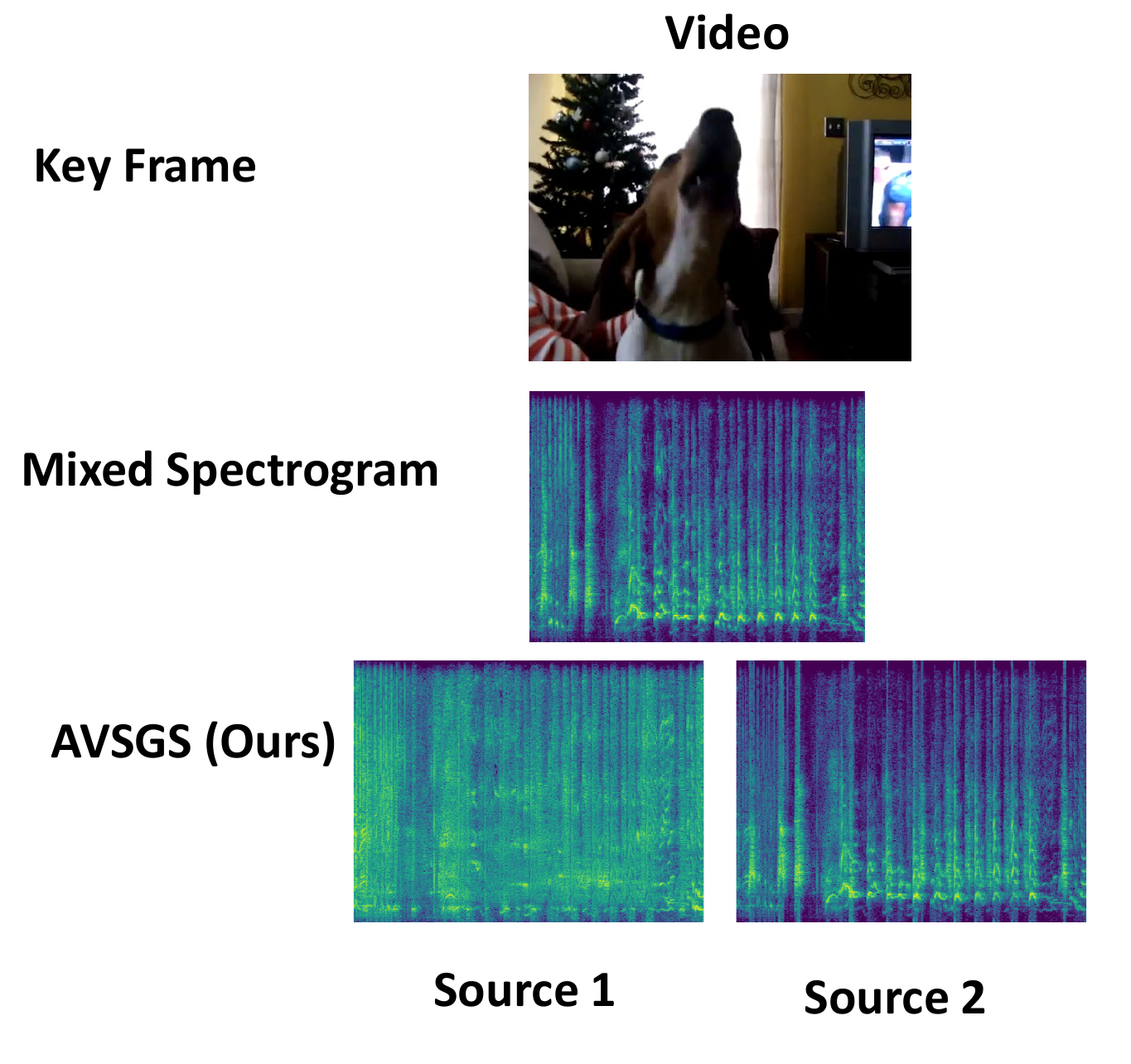} 
   \caption{ Qualitative separation results on a video with 2 sound sources for the ASIW videos. Sample key frame is shown for the videos are shown. The spectrogram of the separated audio is plotted.}
    \label{fig:asiw_perf_duet1}
\end{figure}

\begin{figure}[t]
    \centering
    \includegraphics[width=0.99\columnwidth,trim={0cm 6cm 20cm 0cm},clip=True]{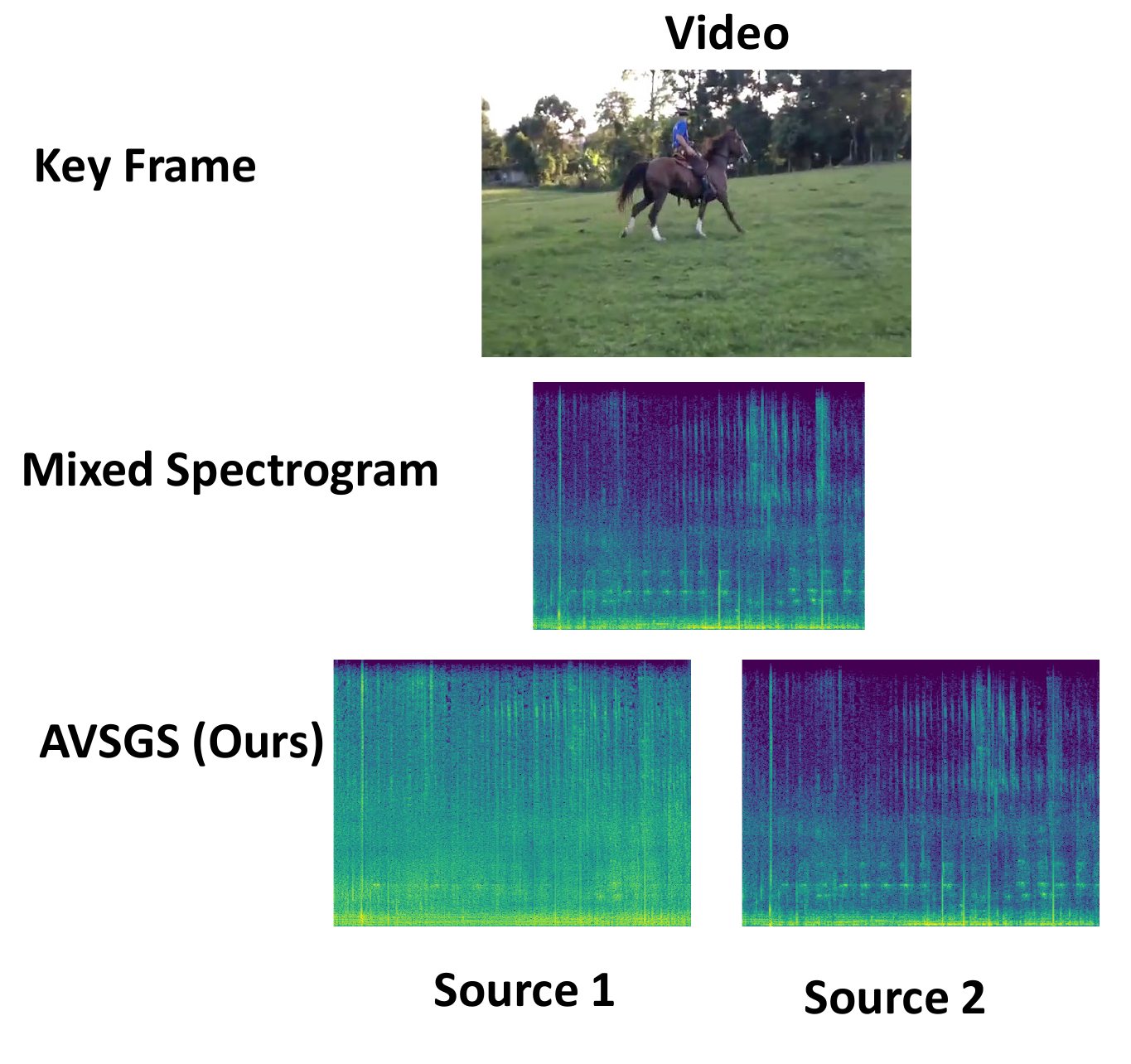} 
   \caption{ Qualitative separation results on a video with 2 sound sources for the ASIW videos. Sample key frame is shown for the videos are shown. The spectrogram of the separated audio is plotted.}
    \label{fig:asiw_perf_duet2}
\end{figure}

\begin{figure}[t]
    \centering
    \includegraphics[width=0.99\columnwidth,trim={0cm 6cm 20cm 0cm},clip=True]{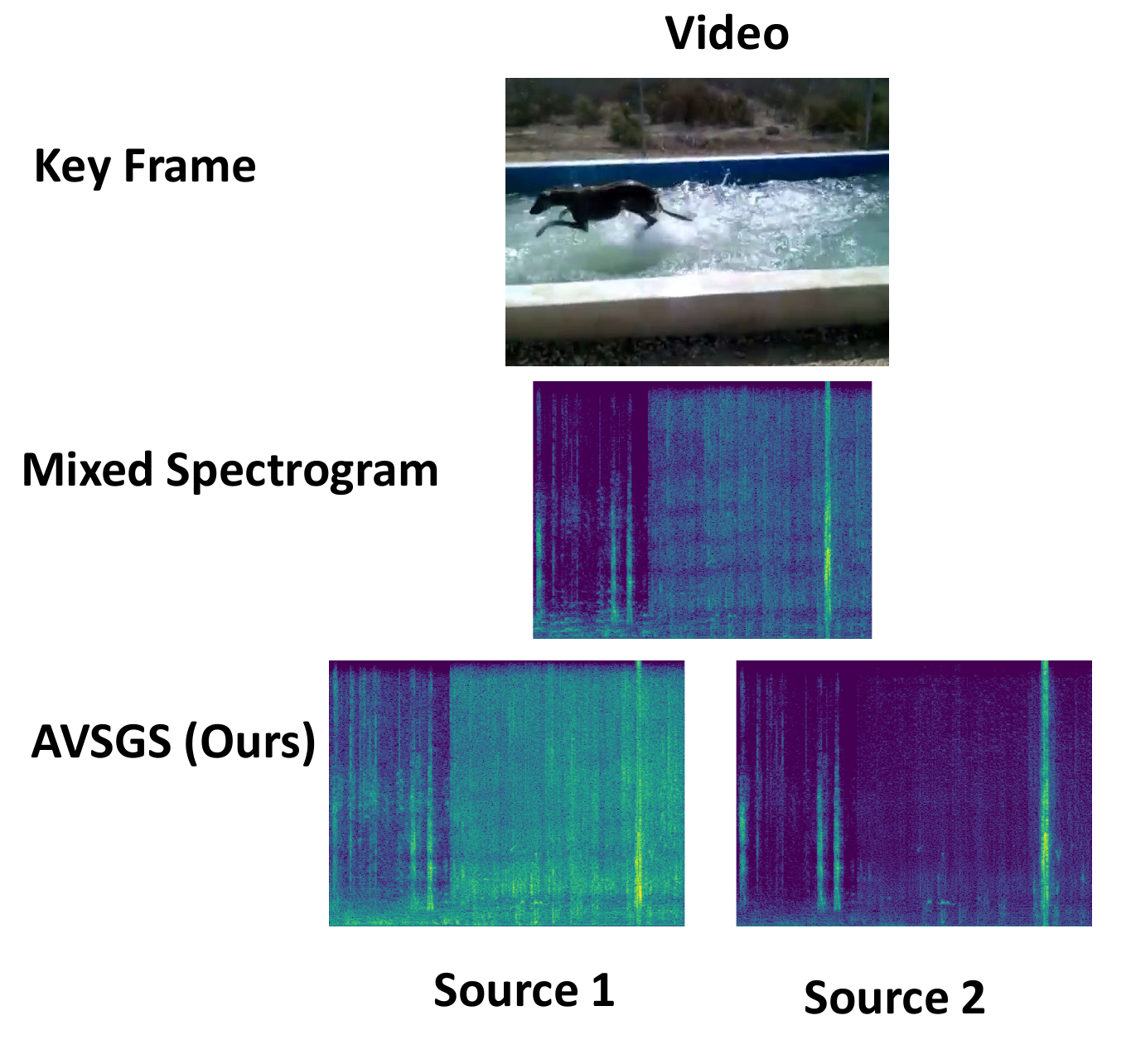} 
   \caption{ Qualitative separation results on a video with 2 sound sources for the ASIW videos. Sample key frame is shown for the videos are shown. The spectrogram of the separated audio is plotted.}
    \label{fig:asiw_perf_duet3}
\end{figure}

\begin{figure}[t]
    \centering
    \includegraphics[width=0.99\columnwidth,trim={0cm 6cm 20cm 0cm},clip=True]{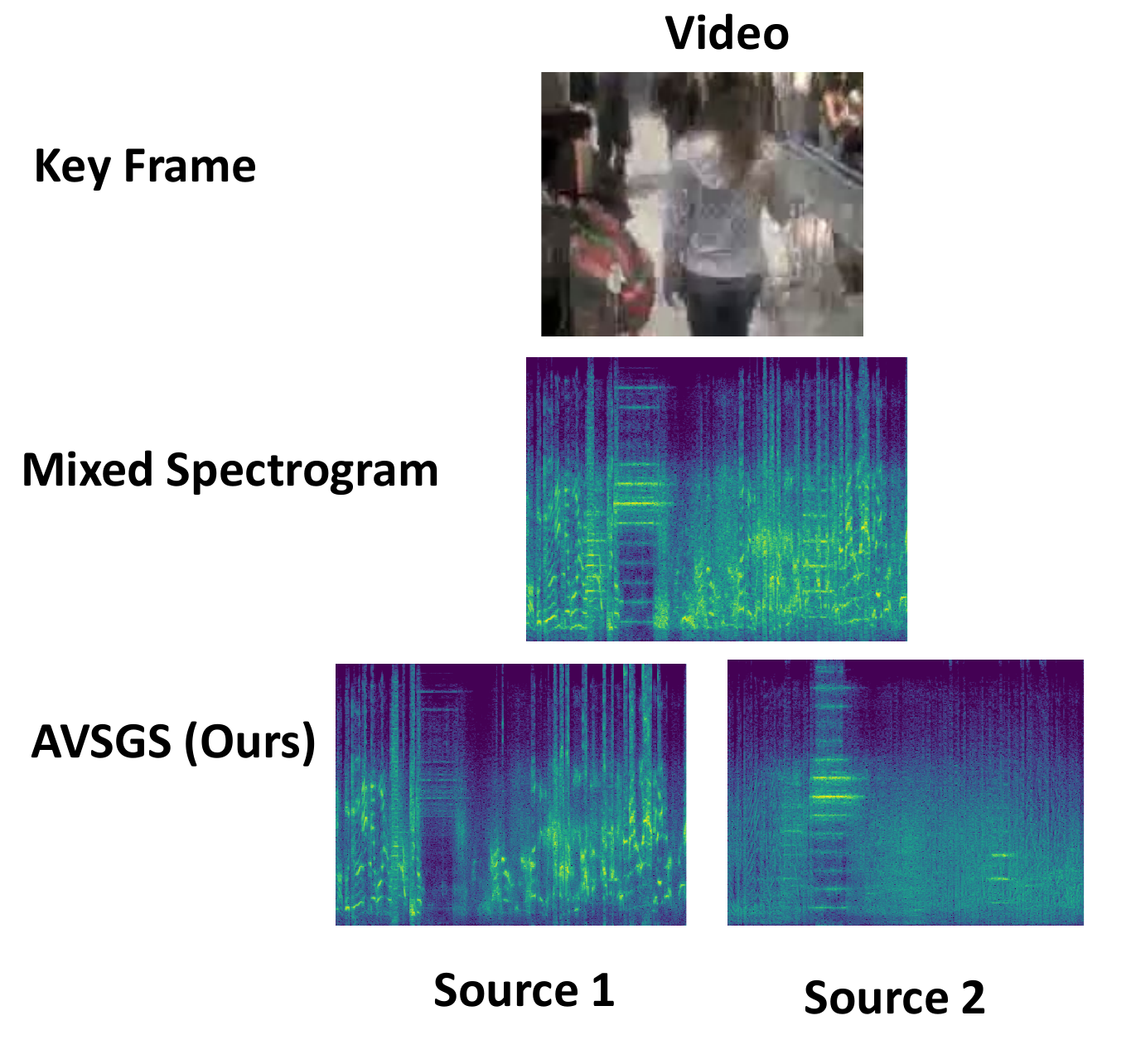} 
   \caption{ Qualitative separation results on a video with 2 sound sources for the ASIW videos. Sample key frame is shown for the videos are shown. The spectrogram of the separated audio is plotted.}
    \label{fig:asiw_perf_duet4}
\end{figure}

\begin{figure}[t]
    \centering
    \includegraphics[width=0.99\columnwidth,trim={0cm 0cm 20cm 0cm},clip=True]{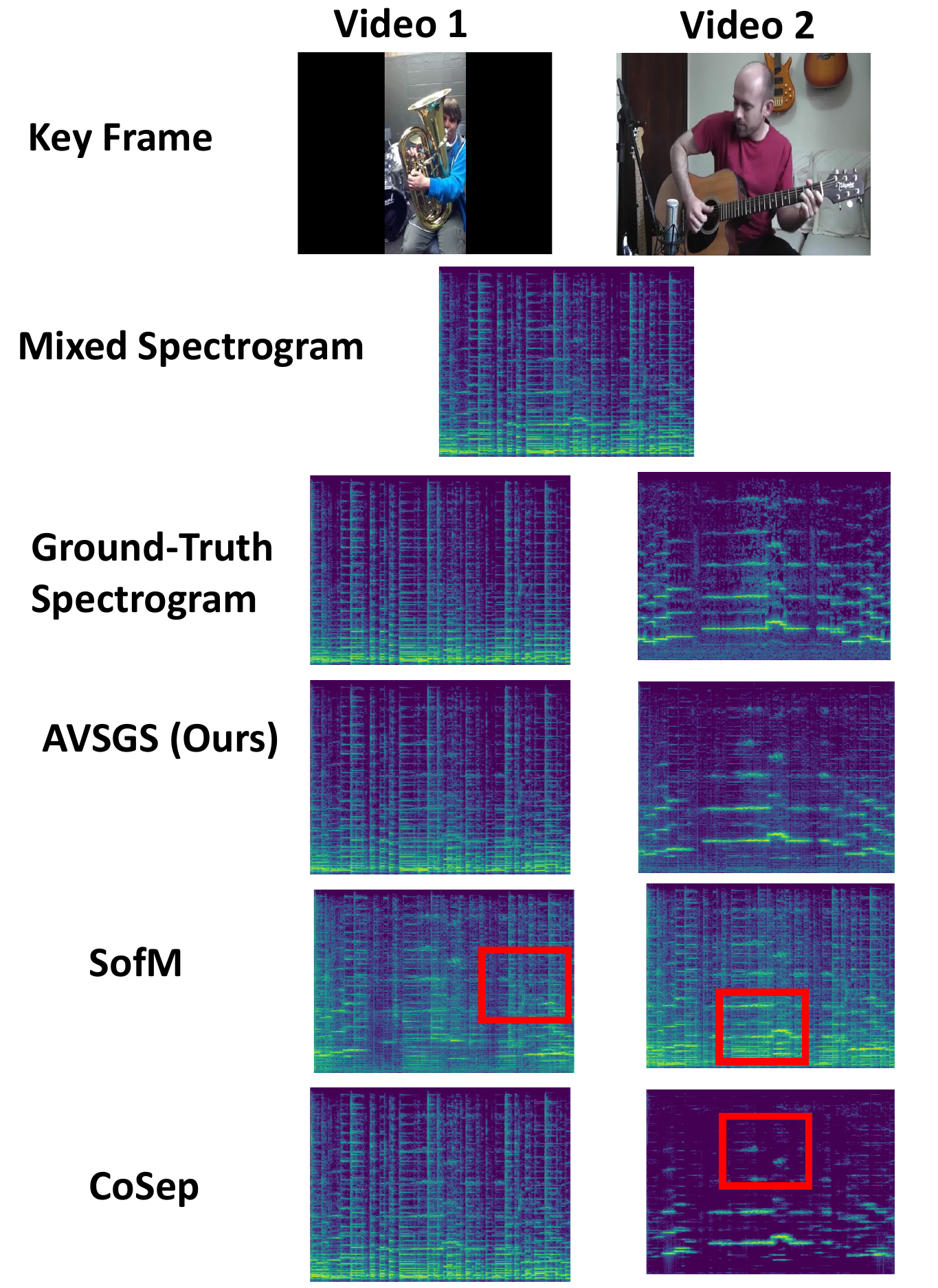} 
   \caption{ Qualitative separation results on a mixture of two MUSIC videos. Sample key frames for both videos are shown. The spectrogram of the mixed audio is plotted as well. Also shown are the separated spectrograms obtained by different methods. Red boxes indicate regions of high differences between ground truth and predicted spectrograms.}
    \label{fig:music_perf_pair1}
\end{figure}

\begin{figure}[t]
    \centering
    \includegraphics[width=0.99\columnwidth,trim={0cm 0cm 20cm 0cm},clip=True]{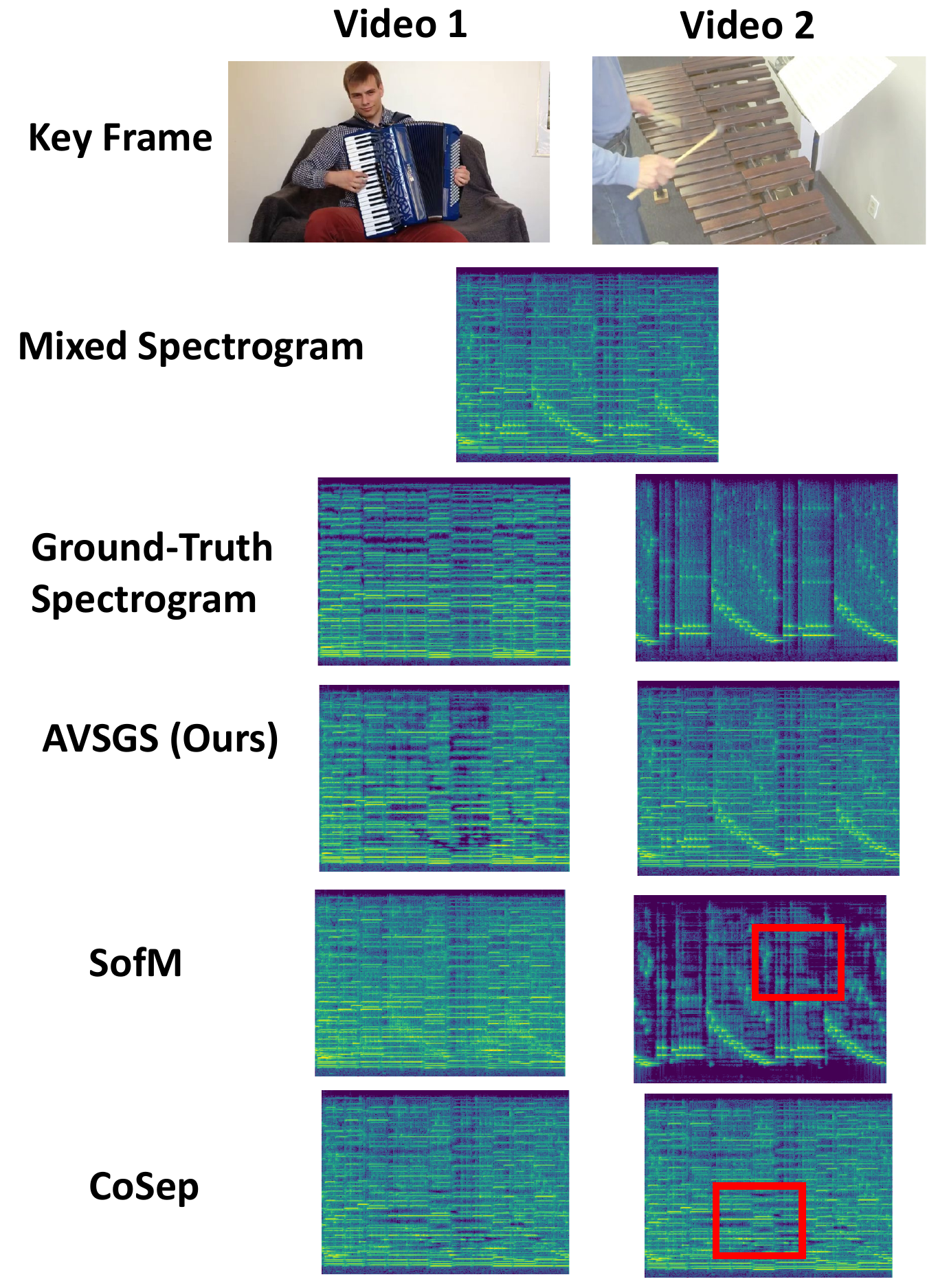} 
   \caption{ Qualitative separation results on a mixture of two MUSIC videos. Sample key frames for both videos are shown. The spectrogram of the mixed audio is plotted as well. Also shown are the separated spectrograms obtained by different methods. Red boxes indicate regions of high differences between ground truth and predicted spectrograms.}
    \label{fig:music_perf_pair2}
\end{figure}

\begin{figure}[t]
    \centering
    \includegraphics[width=0.99\columnwidth,trim={0cm 0cm 20cm 0cm},clip=True]{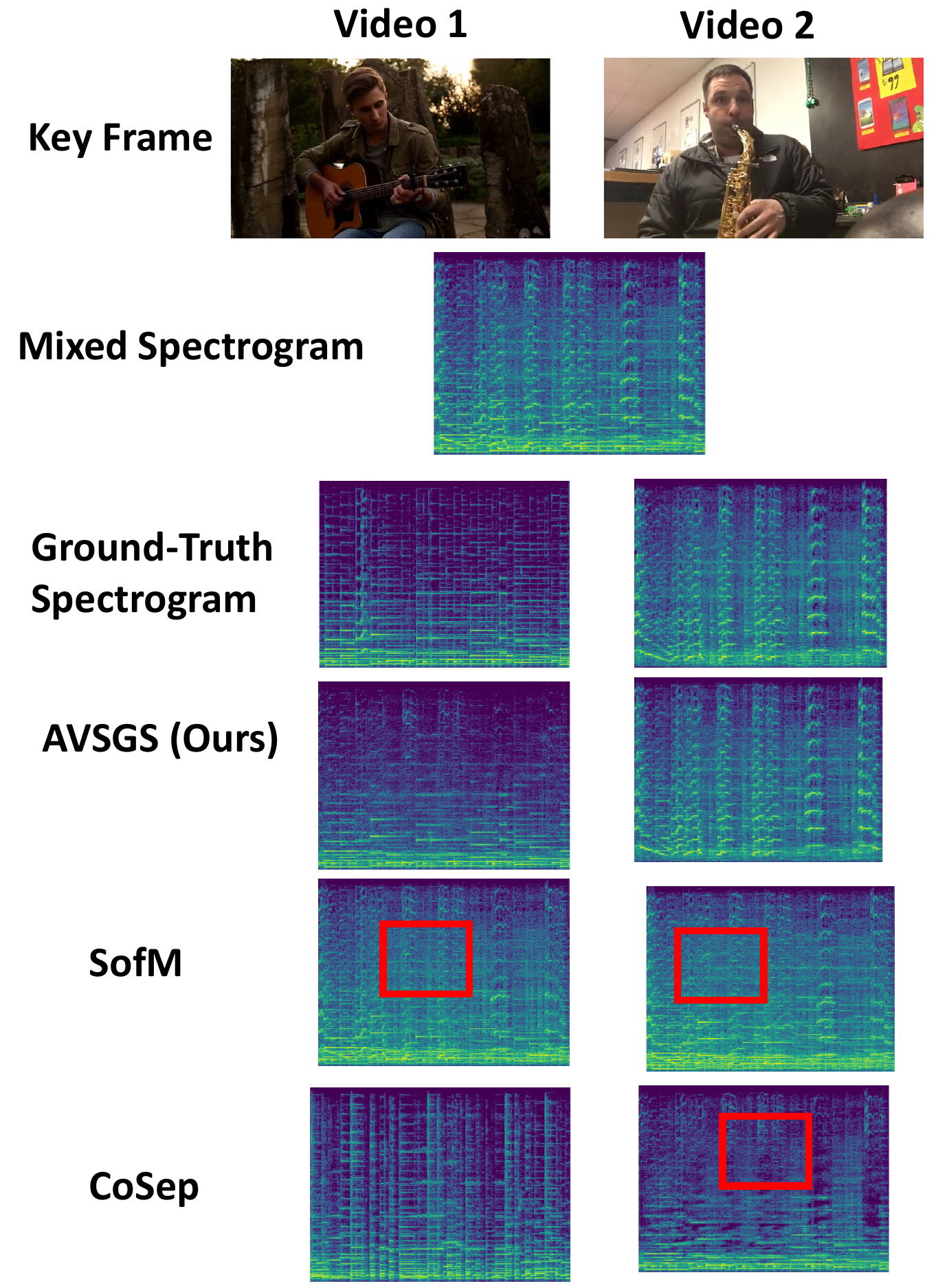} 
   \caption{ Qualitative separation results on a mixture of two MUSIC videos. Sample key frames for both videos are shown. The spectrogram of the mixed audio is plotted as well. Also shown are the separated spectrograms obtained by different methods. Red boxes indicate regions of high differences between ground truth and predicted spectrograms.}
    \label{fig:music_perf_pair3}
\end{figure}

\begin{figure}[t]
    \centering
    \includegraphics[width=0.99\columnwidth,trim={0cm 0cm 20cm 0cm},clip=True]{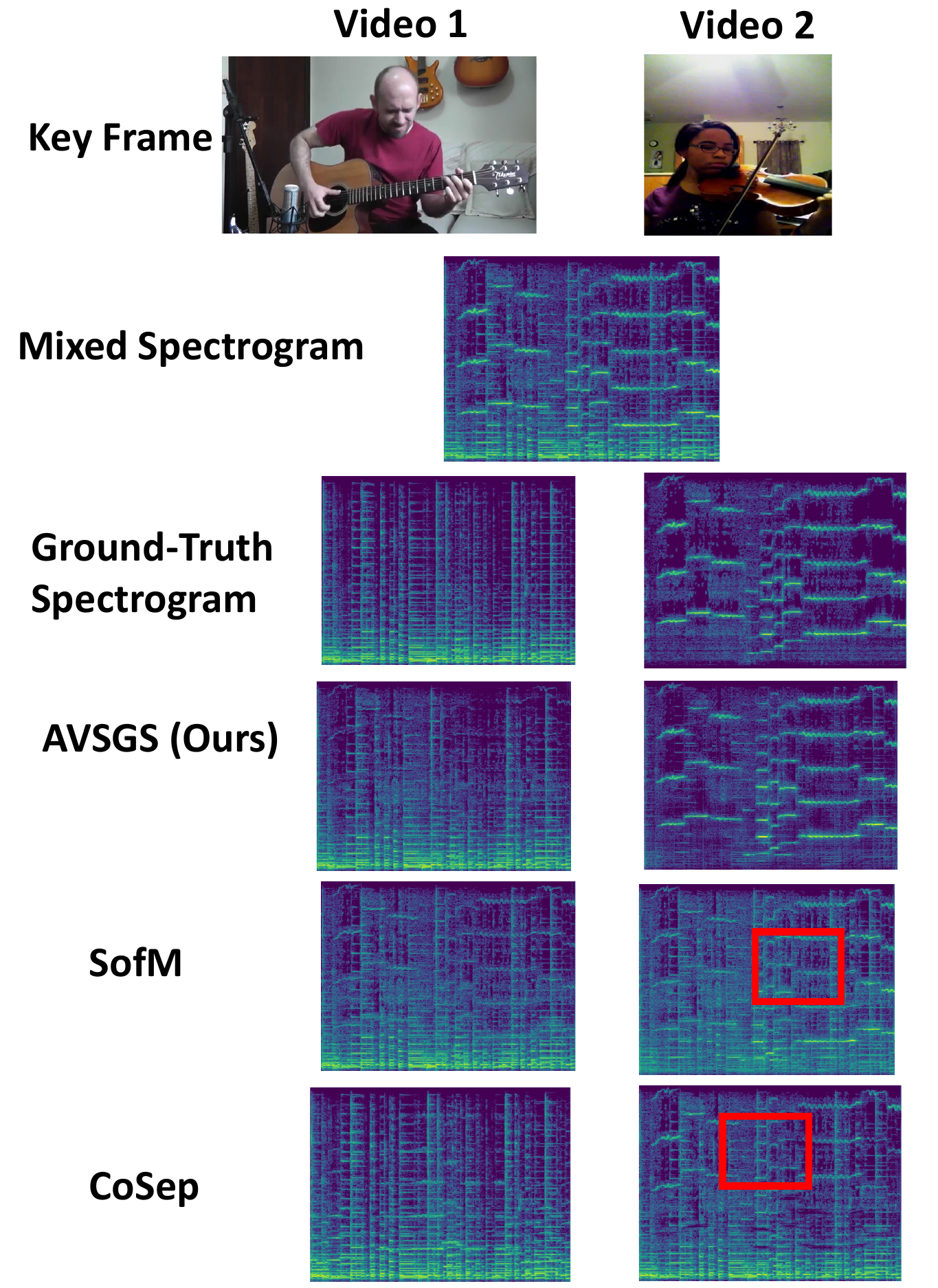} 
   \caption{ Qualitative separation results on a mixture of two MUSIC videos. Sample key frames for both videos are shown. The spectrogram of the mixed audio is plotted as well. Also shown are the separated spectrograms obtained by different methods. Red boxes indicate regions of high differences between ground truth and predicted spectrograms.}
    \label{fig:music_perf_pair4}
\end{figure}

\begin{figure}[t]
    \centering
    \includegraphics[width=0.99\columnwidth,trim={0cm 0cm 20cm 0cm},clip=True]{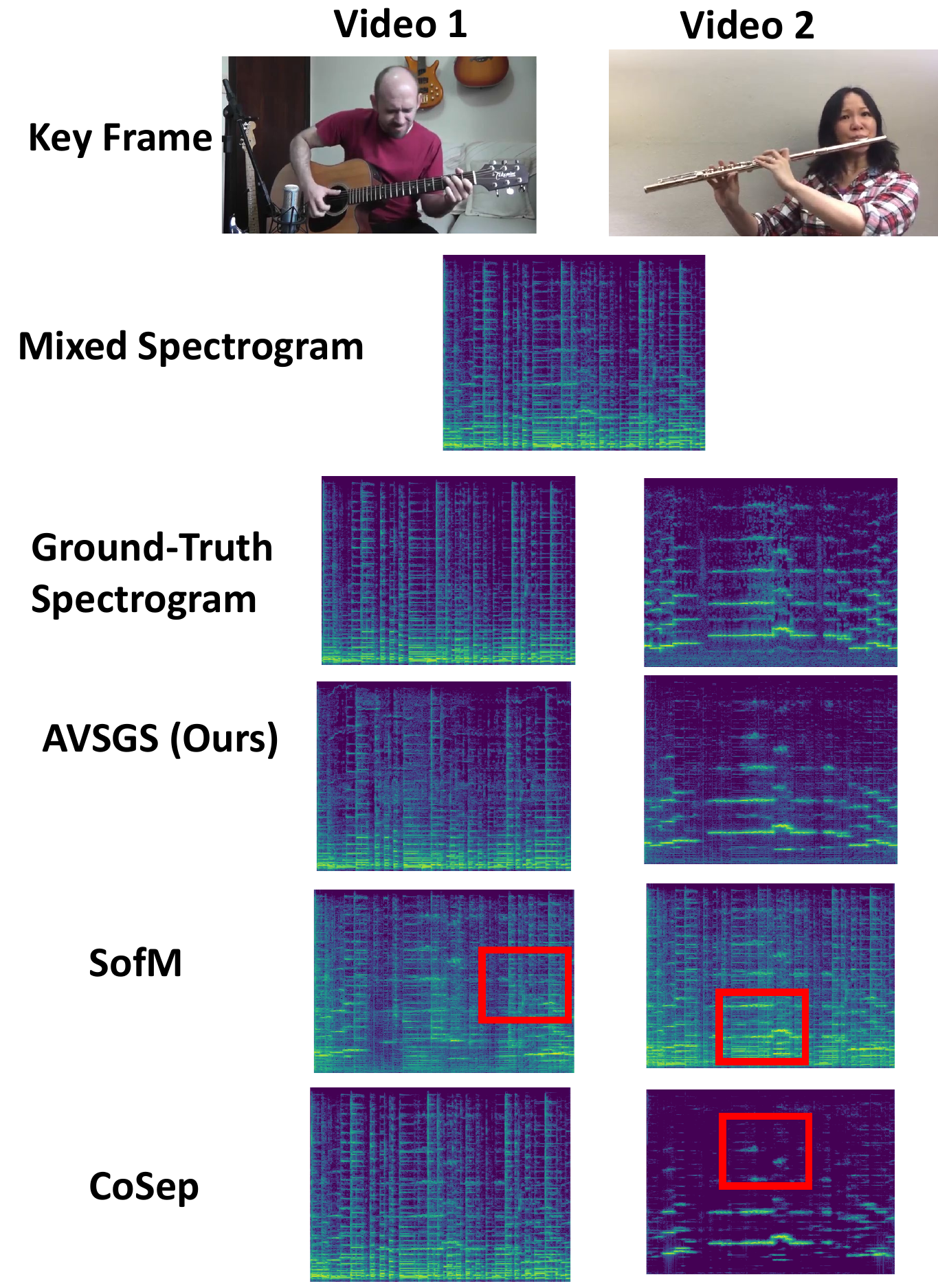} 
   \caption{ Qualitative separation results on a mixture of two MUSIC videos. Sample key frames for both videos are shown. The spectrogram of the mixed audio is plotted as well. Also shown are the separated spectrograms obtained by different methods. Red boxes indicate regions of high differences between ground truth and predicted spectrograms.}
    \label{fig:music_perf_pair5}
\end{figure}

\begin{figure}[t]
    \centering
    \includegraphics[width=0.99\columnwidth,trim={0cm 6cm 20cm 0cm},clip=True]{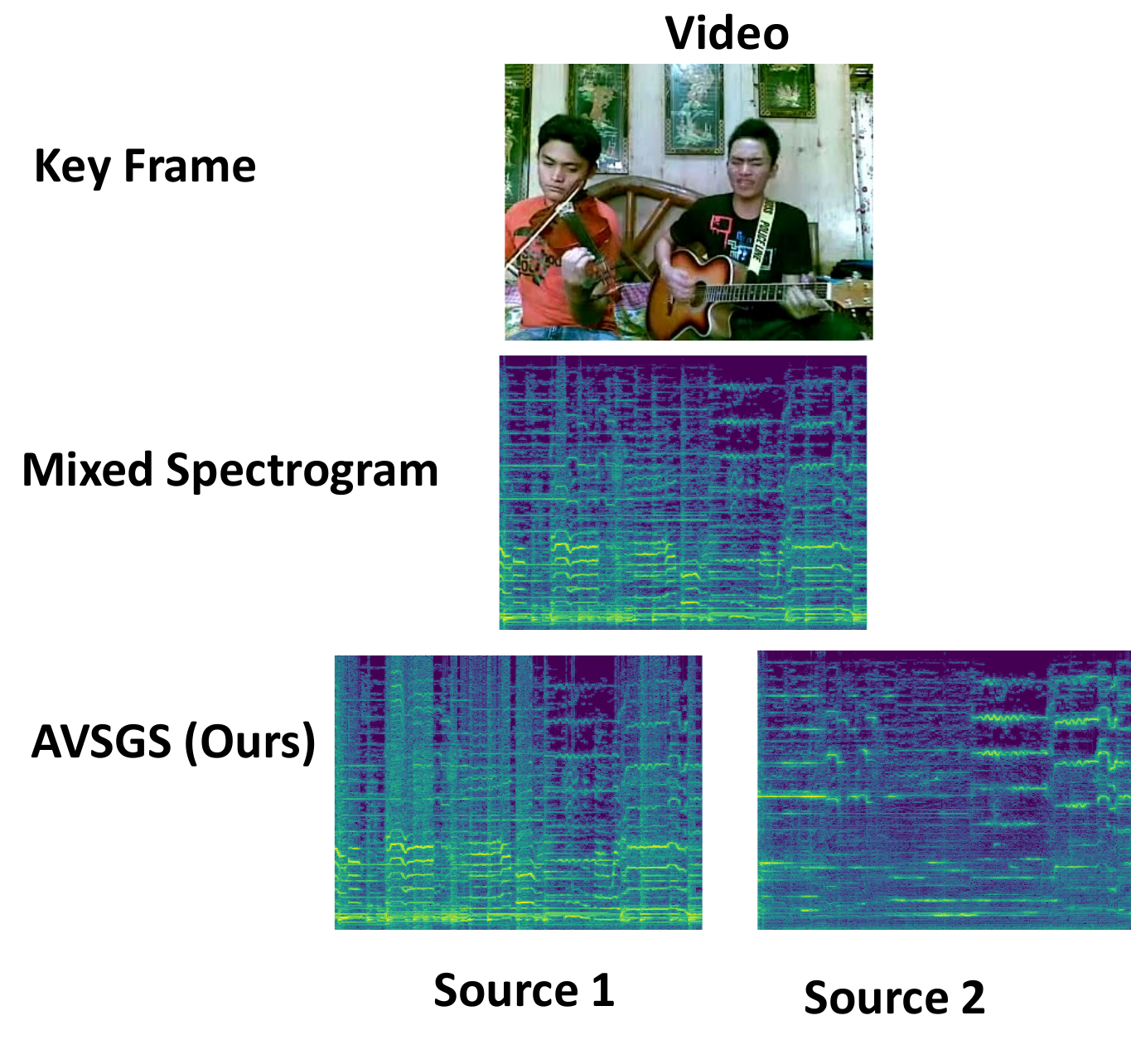} 
   \caption{ Qualitative separation results on a video with 2 sound sources for the MUSIC videos. Sample key frame is shown for the videos are shown. The spectrogram of the separated audio is plotted.}
    \label{fig:music_perf_duet1}
\end{figure}

\begin{figure}[t]
    \centering
    \includegraphics[width=0.99\columnwidth,trim={0cm 6cm 20cm 0cm},clip=True]{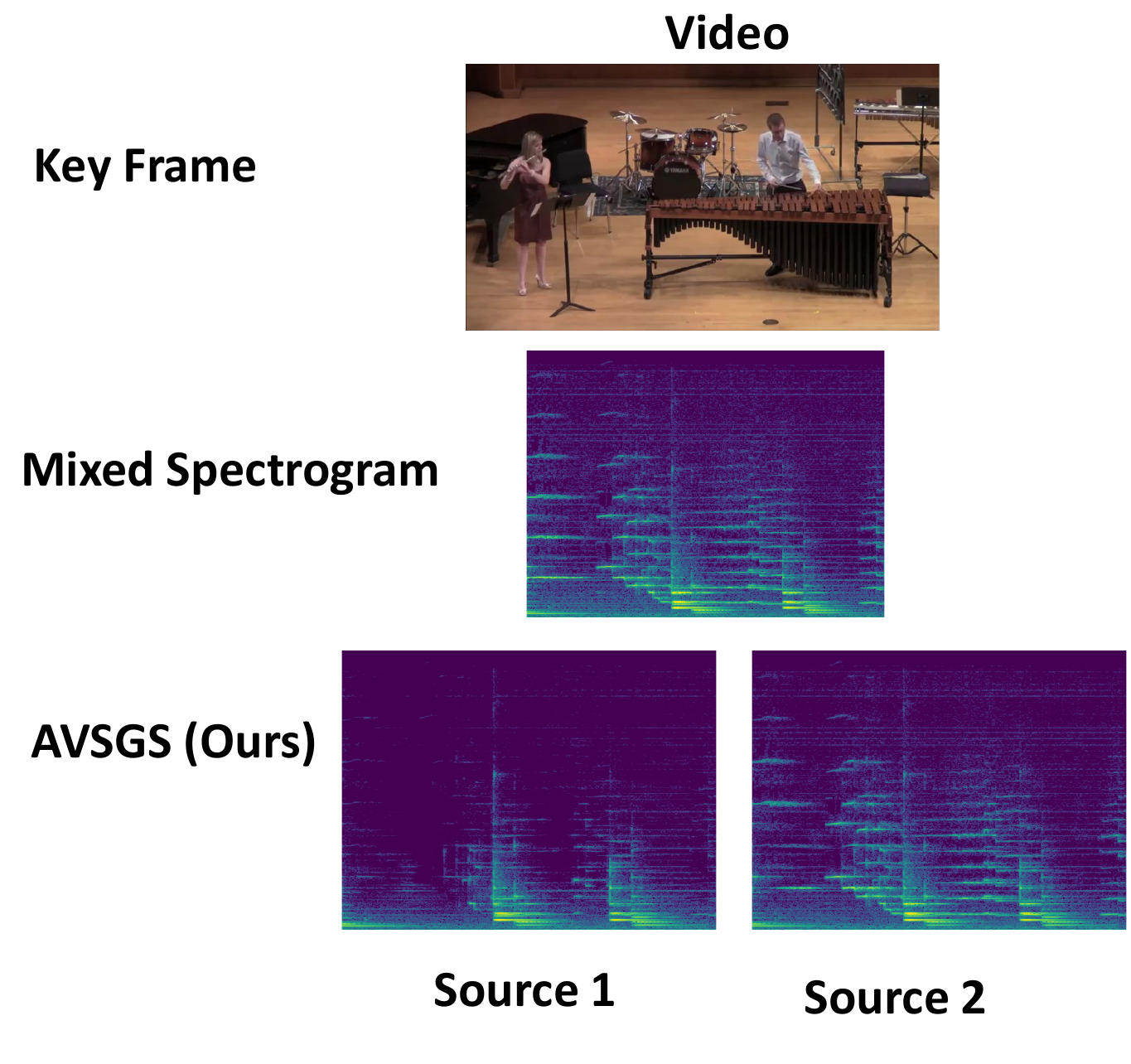} 
   \caption{ Qualitative separation results on a video with 2 sound sources for the MUSIC videos. Sample key frame is shown for the videos are shown. The spectrogram of the separated audio is plotted.}
    \label{fig:music_perf_duet2}
\end{figure}

From the qualitative visualizations presented in Figures~\ref{fig:asiw_perf_pair1},~\ref{fig:asiw_perf_pair2},~\ref{fig:asiw_perf_pair3},~\ref{fig:asiw_perf_pair4},~\ref{fig:asiw_perf_pair5},~\ref{fig:music_perf_pair1},~\ref{fig:music_perf_pair2},~\ref{fig:music_perf_pair3},~\ref{fig:music_perf_pair4},~\ref{fig:music_perf_pair5} we see that AVSGS is better able to separate the audio compared to competing baseline methods on ASIW and MUSIC respectively. We also notice that the separations obtained by AVSGS are more artifact free. Addtionally, in Figures~\ref{fig:asiw_perf_duet1},~\ref{fig:asiw_perf_duet2},~\ref{fig:asiw_perf_duet3},~\ref{fig:asiw_perf_duet4},~\ref{fig:music_perf_duet1},~\ref{fig:music_perf_duet2} we notice that AVSGS is adept at separating multiple sound sources from the same video, as reflected by the difference in the resultant separated spectrograms from the 2 sources.

\subsection{Human Preference Evaluations} 
In order to subjectively assess the quality of audio source separation, we evaluated a randomly chosen subset of separated audio samples from AVSGS and our closest non-MUSIC-specific competitor SofM for human preferability, on both ASIW and MUSIC datasets. Table~\ref{tab:human_eval} reports these results and shows a clear preference of the evaluators, for our method over SofM on average 80-90\% of the time.

\clearpage
\section{List of Auditory Words, Principal Objects, and Frequency in the ASIW Dataset}
\label{sec:listofwords}
\begin{enumerate}
\item \textbf{	babble:}	baby/child/little girl		45
\item \textbf{	babbling:}	baby/child/little girl		8
\item \textbf{	cry:}	baby/child/little girl		1363
\item \textbf{	crying:}	baby/child/little girl		160
\item \textbf{	fidget:}	baby/child/little girl		9
\item \textbf{	giggling:}	baby/child/little girl		6
\item \textbf{	jabbering:}	baby/child/little girl		1
\item \textbf{	singling:}	baby/child/little girl		1
\item \textbf{	sobbing:}	baby/child/little girl		9
\item \textbf{	sobs:}	baby/child/little girl		12
\item \textbf{	spitting:}	baby/child/little girl		2
\item \textbf{	chiming:}	bell		44
\item \textbf{	resonating:}	bell		1
\item \textbf{	rhythmically:}	bell		47
\item \textbf{	warning:}	bell		59
\item \textbf{	calling:}	bird/birds/duck/ducks		10
\item \textbf{	cheep:}	bird/birds/duck/ducks		15
\item \textbf{	chipping:}	bird/birds/duck/ducks		4
\item \textbf{	chirp:}	bird/birds/duck/ducks		2274
\item \textbf{	chirping:}	bird/birds/duck/ducks		53
\item \textbf{	flapping:}	bird/birds/duck/ducks		14
\item \textbf{	flutter:}	bird/birds/duck/ducks		35
\item \textbf{	gobbling:}	bird/birds/duck/ducks		1
\item \textbf{	quacking:}	bird/birds/duck/ducks		104
\item \textbf{	quaking:}	bird/birds/duck/ducks		7
\item \textbf{	squawk:}	bird/birds/duck/ducks		73
\item \textbf{	squawking:}	bird/birds/duck/ducks		4
\item \textbf{	vocalize:}	bird/birds/duck/ducks		193
\item \textbf{	whistling:}	bird/birds/duck/ducks		100
\item \textbf{	click:}	camera		913
\item \textbf{	donging:}	clock/clocks/clock tower/alarm clocks		1
\item \textbf{	locking:}	clock/clocks/clock tower/alarm clocks		2
\item \textbf{	tick:}	clock/clocks/clock tower/alarm clocks		468
\item \textbf{	ticking:}	clock/clocks/clock tower/alarm clocks		187
\item \textbf{	barking:}	dog/dogs		591
\item \textbf{	barks:}	dog/dogs		1
\item \textbf{	growl:}	dog/dogs		305
\item \textbf{	grumbling:}	dog/dogs		2
\item \textbf{	howl:}	dog/dogs		127
\item \textbf{	oinking:}	dog/dogs		119
\item \textbf{	panting:}	dog/dogs		45
\item \textbf{	playfully:}	dog/dogs		10
\item \textbf{	responding:}	dog/dogs		3
\item \textbf{	shakes:}	dog/dogs		1
\item \textbf{	whine:}	dog/dogs		196
\item \textbf{	yap:}	dog/dogs		7
\item \textbf{	emptying:}	drain/toilet/toilet seat/toilet bowl		1
\item \textbf{	flush:}	drain/toilet/toilet seat/toilet bowl		824
\item \textbf{	flushing:}	drain/toilet/toilet seat/toilet bowl		13
\item \textbf{	cantering:}	horse/horses		1
\item \textbf{	clop:}	horse/horses		313
\item \textbf{	clopping:}	horse/horses		33
\item \textbf{	galloping:}	horse/horses		18
\item \textbf{	neighs:}	horse/horses		2
\item \textbf{	oping:}	horse/horses		1
\item \textbf{	riding:}	horse/horses		4
\item \textbf{ oping:} horse/horses		1
\item \textbf{	trotting:}	horse/horses		12
\item \textbf{	achoo:}	man/woman/young man/people		1
\item \textbf{	amplified:}	man/woman/young man/people		9
\item \textbf{	applaud:}	man/woman/young man/people		289
\item \textbf{	applauding:}	man/woman/young man/people		22
\item \textbf{	appreciatively:}	man/woman/young man/people		1
\item \textbf{	articulately:}	man/woman/young man/people		1
\item \textbf{	breathing:}	man/woman/young man/people		132
\item \textbf{	burp:}	man/woman/young man/people		267
\item \textbf{	celebrate:}	man/woman/young man/people		2
\item \textbf{	chant:}	man/woman/young man/people		50
\item \textbf{	chanting:}	man/woman/young man/people		2
\item \textbf{	cheer:}	man/woman/young man/people		623
\item \textbf{	cheering:}	man/woman/young man/people		103
\item \textbf{	chuckle:}	man/woman/young man/people		98
\item \textbf{	clapping:}	man/woman/young man/people		40
\item \textbf{	communicating:}	man/woman/young man/people		6
\item \textbf{	conversation:}	man/woman/young man/people		158
\item \textbf{	converse:}	man/woman/young man/people		91
\item \textbf{	coughing:}	man/woman/young man/people		21
\item \textbf{	coughs:}	man/woman/young man/people		1
\item \textbf{	crunching:}	man/woman/young man/people		33
\item \textbf{	curtly:}	man/woman/young man/people		1
\item \textbf{	dialog:}	man/woman/young man/people		4
\item \textbf{	echo:}	man/woman/young man/people		141
\item \textbf{	eruption:}	man/woman/young man/people		3
\item \textbf{	exhaling:}	man/woman/young man/people		1
\item \textbf{	falsetto:}	man/woman/young man/people		1
\item \textbf{	fighting:}	man/woman/young man/people		3
\item \textbf{	flicking:}	man/woman/young man/people		1
\item \textbf{	folding:}	man/woman/young man/people		1
\item \textbf{	forklift:}	man/woman/young man/people		1
\item \textbf{	gag:}	man/woman/young man/people		6
\item \textbf{	girlish:}	man/woman/young man/people		1
\item \textbf{	glee:}	man/woman/young man/people		1
\item \textbf{	hoots:}	man/woman/young man/people		1
\item \textbf{	indistinctly:}	man/woman/young man/people		7
\item \textbf{	inhale:}	man/woman/young man/people		20
\item \textbf{	kaboom:}	man/woman/young man/people		1
\item \textbf{	laugh:}	man/woman/young man/people		3091
\item \textbf{	laughing:}	man/woman/young man/people		270
\item \textbf{	manspaking:}	man/woman/young man/people		1
\item \textbf{	melody:}	man/woman/young man/people		24
\item \textbf{	moaning:}	man/woman/young man/people		1
\item \textbf{	monotone:}	man/woman/young man/people		10
\item \textbf{	murmur:}	man/woman/young man/people		91
\item \textbf{	narrating:}	man/woman/young man/people		11
\item \textbf{	playing:}	man/woman/young man/people		123
\item \textbf{	prancing:}	man/woman/young man/people		1
\item \textbf{	recording:}	man/woman/young man/people		8
\item \textbf{	reverberate:}	man/woman/young man/people		9
\item \textbf{	reverberating:}	man/woman/young man/people		4
\item \textbf{	screaming:}	man/woman/young man/people		32
\item \textbf{	scuffling:}	man/woman/young man/people		4
\item \textbf{	sigh:}	man/woman/young man/people		39
\item \textbf{	sighing:}	man/woman/young man/people		2
\item \textbf{	slurp:}	man/woman/young man/people		10
\item \textbf{	slurping:}	man/woman/young man/people		1
\item \textbf{	sneezing:}	man/woman/young man/people		24
\item \textbf{	sniffing:}	man/woman/young man/people		7
\item \textbf{	sniveling:}	man/woman/young man/people		1
\item \textbf{	snort:}	man/woman/young man/people		65
\item \textbf{	stuttering:}	man/woman/young man/people		4
\item \textbf{	subdued:}	man/woman/young man/people		5
\item \textbf{	thumping:}	man/woman/young man/people		123
\item \textbf{	thunderous:}	man/woman/young man/people		5
\item \textbf{	uproar:}	man/woman/young man/people		4
\item \textbf{	uproarious:}	man/woman/young man/people		1
\item \textbf{	uproariously:}	man/woman/young man/people		1
\item \textbf{	verbally:}	man/woman/young man/people		2
\item \textbf{	vigorously:}	water/water tank/water bottle		17
\item \textbf{	yelling:}	man/woman/young man/people		74
\item \textbf{	yodel:}	man/woman/young man/people		1
\item \textbf{	baaing:}	sheep/goat/goats/chicken		114
\item \textbf{	bleat:}	sheep/goat/goats/chicken		583
\item \textbf{	cackle:}	sheep/goat/goats/chicken		13
\item \textbf{	answering:}	telephone		4
\item \textbf{	ringing:}	telephone		218
\item \textbf{	chug:}	train/trains/train car/train cars/passenger train/train engine		133
\item \textbf{	sounding:}	train/trains/train car/train cars/passenger train/train engine		8
\item \textbf{	backing:}	vehicle/car/cars/truck/trucks		2
\item \textbf{	beeps:}	vehicle/car/cars/truck/trucks		2
\item \textbf{	brake:}	vehicle/car/cars/truck/trucks		76
\item \textbf{	braking:}	vehicle/car/cars/truck/trucks		2
\item \textbf{	breaks:}	vehicle/car/cars/truck/trucks		1
\item \textbf{	driving:}	vehicle/car/cars/truck/trucks		25
\item \textbf{	honk:}	vehicle/car/cars/truck/trucks		584
\item \textbf{	racing:}	vehicle/car/cars/truck/trucks		50
\item \textbf{	raggedly:}	vehicle/car/cars/truck/trucks		1
\item \textbf{	roving:}	vehicle/car/cars/truck/trucks		2
\item \textbf{	shifting:}	vehicle/car/cars/truck/trucks		16
\item \textbf{	silently:}	vehicle/car/cars/truck/trucks		3
\item \textbf{	skidding:}	vehicle/car/cars/truck/trucks		15
\item \textbf{	draining:}	water/water tank/water bottle		1
\item \textbf{	drip:}	water/water tank/water bottle		106
\item \textbf{	flowing:}	water/water tank/water bottle		9
\item \textbf{	gushing:}	water/water tank/water bottle		1
\item \textbf{	hisses:}	water/water tank/water bottle		1
\item \textbf{	jostling:}	water/water tank/water bottle		2
\item \textbf{	leaking:}	water/water tank/water bottle		1
\item \textbf{	pouring:}	water/water tank/water bottle		14
\item \textbf{	raining:}	water/water tank/water bottle		6
\item \textbf{	splashing:}	water/water tank/water bottle		211
\item \textbf{	splay:}	water/water tank/water bottle		1
\item \textbf{	trickling:}	water/water tank/water bottle		22
\item \textbf{	woosh:}	water/water tank/water bottle		3
\item \textbf{	audible:}	background		22
\item \textbf{	audibly:}	background		1
\item \textbf{	banging:}	background		191
\item \textbf{	beat:}	background		53
\item \textbf{	beatable:}	background		1
\item \textbf{	beating:}	background		5
\item \textbf{	beep:}	background		910
\item \textbf{	bellow:}	background		2
\item \textbf{	blast:}	background		79
\item \textbf{	blowing:}	background	183
\item \textbf{	boiling:}	background		1
\item \textbf{	bouncing:}	background		6
\item \textbf{	brushing:}	background		4
\item \textbf{	buffeting:}	background		3
\item \textbf{	bumble:}	background		1
\item \textbf{	burble:}	background		31
\item \textbf{	burbling:}	background		1
\item \textbf{	burning:}	background		4
\item \textbf{	bursting:}	background		6
\item \textbf{	buzzer:}	background		13
\item \textbf{	chang:}	background		1
\item \textbf{	chewing:}	background		4
\item \textbf{	chocking:}	background		1
\item \textbf{	choke:}	background	m	6
\item \textbf{	churning:}	background		3
\item \textbf{	clacking:}	background		173
\item \textbf{	clang:}	background		197
\item \textbf{	clank:}	background		597
\item \textbf{	clanking:}	background		334
\item \textbf{	clattering:}	background		39
\item \textbf{	clinking:}	background		76
\item \textbf{	clumping:}	background		1
\item \textbf{	clunking:}	background		8
\item \textbf{	cluttering:}	background		2
\item \textbf{	cocking:}	background		9
\item \textbf{	collision:}	background		3
\item \textbf{	crack:}	background		126
\item \textbf{	cracking:}	background		24
\item \textbf{	cranking:}	background		13
\item \textbf{	crinkling:}	background		103
\item \textbf{	croak:}	background		339
\item \textbf{	croaking:}	background		22
\item \textbf{	crumpling:}	background		63
\item \textbf{	dabbling:}	background		1
\item \textbf{	deafen:}	background		1
\item \textbf{	dinging:}	background		2
\item \textbf{	drooping:}	background		1
\item \textbf{	explode:}	background		53
\item \textbf{	fainting:}	background		1
\item \textbf{	faintly:}	background		253
\item \textbf{	filing:}	background		20
\item \textbf{	firing:}	background		35
\item \textbf{	fizzing:}	background		1
\item \textbf{	flipping:}	background		3
\item \textbf{	fumbling:}	background		1
\item \textbf{	grinding:}	background		34
\item \textbf{	grunting:}	background		28
\item \textbf{	gulping	:} background		2
\item \textbf{	gusting:}	background		4
\item \textbf{	heaving:}	background		1
\item \textbf{	hoovering:}	background		1
\item \textbf{	hovering:}	background		8
\item \textbf{	humming:}	background		639
\item \textbf{	jarring:}	background		1
\item \textbf{	jumble:}	background		1
\item \textbf{	launching:}	background		1
\item \textbf{	licking:}	background		1
\item \textbf{	loudly:}	background		1828
\item \textbf{	mingle:}	background		2
\item \textbf{	mix:}	background		22
\item \textbf{	mixer:}	background		1
\item \textbf{	noise:}	background		2529
\item \textbf{	noisily:}	background		13
\item \textbf{	outburst:}	background		3
\item \textbf{	popping:}	background		69
\item \textbf{	pound:}	background		24
\item \textbf{	puffing:}	background		2
\item \textbf{	pulsing:}	background		4
\item \textbf{	rabbiting:}	background		2
\item \textbf{	ragged:}	background		2
\item \textbf{	raging:}	background		1
\item \textbf{	rapping:}	background		6
\item \textbf{	ratcheting:}	background		5
\item \textbf{	rattling:}	background		113
\item \textbf{	reeving:}	background		1
\item \textbf{	releasing:}	background		20
\item \textbf{	reloading:}	background		5
\item \textbf{	revving:}	background		227
\item \textbf{	rewinding:}	background		1
\item \textbf{	rhythm:}	background		18
\item \textbf{	ripping:}	background		12
\item \textbf{	roaring:}	background		71
\item \textbf{	rocking:}	background		1
\item \textbf{	roughly:}	background		34
\item \textbf{	rumbling:}	background		72
\item \textbf{	rustling:}	background		543
\item \textbf{	sanding:}	background		20
\item \textbf{	sawing:}	background		30
\item \textbf{	scowl:}	background		1
\item \textbf{	scrapping:}	background		9
\item \textbf{	shaking:}	background		3
\item \textbf{	sharpen:}	background		3
\item \textbf{	sharpening:}	background		1
\item \textbf{	shrill:}	background		18
\item \textbf{	slashing:}	background		1
\item \textbf{	slightly:}	background		50
\item \textbf{	smash:}	background		6
\item \textbf{	smashing:}	background		4
\item \textbf{	snare:}	background		3
\item \textbf{	snarling:}	background		1
\item \textbf{	sparking:}	background		2
\item \textbf{	splat:}	background		7
\item \textbf{	splattering:}	background		1
\item \textbf{	spraying:}	background		70
\item \textbf{	springing:}	background		1
\item \textbf{	spurt:}	background		4
\item \textbf{	squeaking:}	background		61
\item \textbf{	squealing:}	background		60
\item \textbf{	steadily:}	background		76
\item \textbf{	stitching:}	background		7
\item \textbf{	stretching:}	background		1
\item \textbf{	striking:}	background		2
\item \textbf{	suction:}	background		5
\item \textbf{	swarm:}	background		40
\item \textbf{	swishing:}	background		21
\item \textbf{	tapping:}	background		192
\item \textbf{	thud:}	background		83
\item \textbf{	thudding:}	background		1
\item \textbf{	thwacking:}	background		3
\item \textbf{	tinkling:}	background		8
\item \textbf{	trill:}	background		3
\item \textbf{	tumbling:}	background		2
\item \textbf{	typing:}	background		129
\item \textbf{	vibrantly:}	background		1
\item \textbf{	vibrate:}	background		397
\item \textbf{	vibrating:}	background		66
\item \textbf{	weirdly:}	background		1
\item \textbf{	whirring:}	background		176
\item \textbf{	whooshing:}	background		123
\item \textbf{	winding:}	background		2
\item \textbf{	wishing:}	background		1
\item \textbf{	zapping:}	background		1
\item \textbf{	zipping:}	background		3 
\end{enumerate}

\end{document}